\documentclass[10pt,twocolumn,letterpaper]{article}

\usepackage{cvpr}
\usepackage{times}
\usepackage{epsfig}
\usepackage{graphicx}
\usepackage{amsmath}
\usepackage{amssymb}
\usepackage{indentfirst}
\usepackage{placeins}
\usepackage{pifont}
\usepackage{multirow}
\usepackage{bbding}
\usepackage{bm}
\usepackage{subcaption}
\usepackage[table]{xcolor}

\usepackage[subpreambles=true]{standalone}
\usepackage[utf8]{inputenc}
\usepackage[english]{babel}
\usepackage{import}

\usepackage{phonetic}
\usepackage{cite}
\usepackage{enumitem}
\usepackage[percent]{overpic}
\usepackage{multirow}
\usepackage{url}
\usepackage{xcolor}
\usepackage[linesnumbered]{algorithm2e}

\usepackage[pagebackref=true,breaklinks=true,letterpaper=true,colorlinks,bookmarks=false]{hyperref}

\cvprfinalcopy 

\def\cvprPaperID{7466} 

\newcommand{\Sim}{\mathord{\sim}}
\newcommand{\Paragraph}[1]{\vspace{1.25mm} \noindent \textbf{#1} \hspace{0mm}}
\newcommand{\Section}[1]{\vspace{-1.4mm} \section{#1} \vspace{-1.5mm}}
\newcommand{\SubSection}[1]{\vspace{-1.2mm} \subsection{#1} \vspace{-1.2mm}}
\newcommand{\SubSubSection}[1]{\vspace{-3mm} \subsubsection{#1} \vspace{-1mm}}

\def\figvspace{{\vspace{-4mm}}}
\newcommand{\argmin}{\mathop{\mathrm{argmin}}\limits}
\newcommand\norm[1]{\left\lVert#1\right\rVert}
\begin{document}

\title{The Edge of Depth: Explicit Constraints between Segmentation and Depth}


\author{Shengjie Zhu, Garrick Brazil, Xiaoming Liu \\
Michigan State University, East Lansing MI \\
{\tt\small \{zhusheng, brazilga, liuxm\}@msu.edu}
}

\maketitle

\begin{abstract}
In this work we study the mutual benefits of two common computer vision tasks, self-supervised depth estimation and semantic segmentation from images. 
For example, to help unsupervised monocular depth estimation, constraints from semantic segmentation has been explored implicitly such as sharing and transforming features.
In contrast, we propose to explicitly measure the border consistency between segmentation and depth and minimize it in a greedy manner by iteratively supervising the network towards a locally optimal solution.
Partially this is motivated by our observation that semantic segmentation even trained with limited ground truth ($200$ images of KITTI) can offer more accurate border than that of any (monocular or stereo) image-based depth estimation.
Through extensive experiments, our proposed approach advances the state of the art on unsupervised monocular depth estimation in the KITTI.

\end{abstract}

\Section{Introduction}
Estimating depth is a fundamental problem in computer vision with notable applications in self-driving~\cite{m3d-rpn-monocular-3d-region-proposal-network-for-object-detection} and virtual/augmented reality. 
To solve the challenge, a diverse set of sensors has been utilized ranging from monocular camera~\cite{godard2017unsupervised}, multi-view cameras~\cite{cheng2018learning}, and depth completion from LiDAR~\cite{depth-coefficients-for-depth-completion}. 
Although the monocular system is the least expensive, it is the most challenging due to scale ambiguity.
The current highest performing monocular methods~\cite{yang2018deep, guo2018learning, luo2018single, kuznietsov2017semi, fu2018deep} are reliant on {\it supervised} training, thus consuming large amounts of labelled depth data. 
Recently, {\it self-supervised} methods with photometric supervision have made significant progress by leveraging unlabeled stereo images~\cite{garg2016unsupervised, godard2017unsupervised} or monocular videos~\cite{zhou2017unsupervised, vijayanarasimhan2017sfm, yin2018geonet} to approach comparable performance as the supervised methods.

Yet, self-supervised depth inference techniques suffer from high ambiguity and sensitivity in low-texture regions, reflective surfaces, and the presence of occlusion, likely leading to a sub-optimal solution. 
To reduce these effects, many works seek to incorporate constraints from external modalities. 
For example, prior works have explored leveraging diverse modalities such as optical flow~\cite{yin2018geonet}, surface normal~\cite{yang2017unsupervised}, and semantic segmentation~\cite{chen2019towards, wang2015towards, mousavian2016joint, zhang2019pattern}. 
Optical flow can be naturally linked to depth via ego-motion and object motion, while surface normal can be re-defined as direction of the depth gradient in 3D. 
Comparatively, semantic segmentation is unique in that, though highly relevant, it is difficult to form definite relationship with depth.

\definecolor{darkred}{rgb}{0.65,0.1,0.1}
\definecolor{darkblue}{rgb}{0.1,0.1,0.65}
\definecolor{darkgreen}{rgb}{0.1,0.65,0.1}

\begin{figure}
      \centering
      \includegraphics[width = 0.8\linewidth]{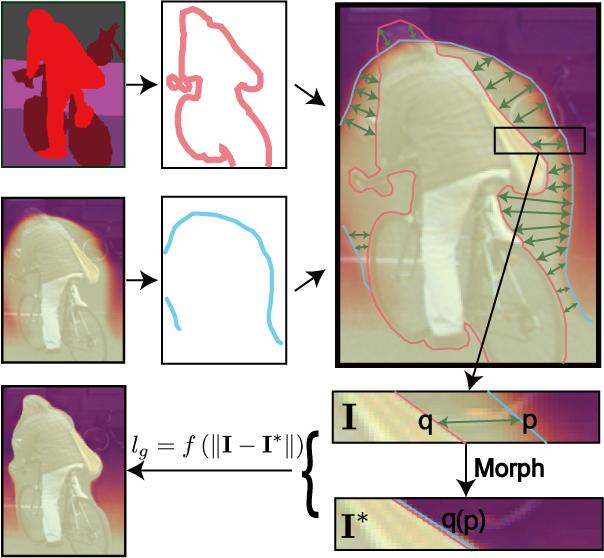}
\vspace{-2mm}
    \caption{
    We explicitly regularize the depth border to be consistent with segmentation border. A ``better" depth $\mathbf{I^*}$ is created through morphing according to distilled point pairs $\mathbf{pq}$. By penalizing its difference with the original prediction $\mathbf{I}$ at each training step, we gradually achieve a more consistent border. The morph happens over every distilled pairs but only one pair illustrated, due to limited space. 
    }
    \label{fig:teaser}
 \figvspace
\end{figure}

In response, prior works tend to model the relation of semantic segmentation and depth {\it implicitly}~\cite{chen2019towards, wang2015towards, mousavian2016joint, zhang2019pattern}. 
For instance,~\cite{chen2019towards, wang2015towards} show that jointly training a shared network with semantic segmentation and depth is helpful to both.\linebreak
\cite{zhang2019pattern} learns a transformation between semantic segmentation and depth feature spaces.
Despite empirically positive results, such techniques lack clear and detailed explanation for their improvement. 
Moreover, prior work has yet to explore the relationship from one of the most obvious aspects --- the shared borders between segmentation and depth.

Hence, we aim to {\it explicitly} constrain monocular self-supervised depth estimation to be more consistent and aligned to its segmentation counterpart.
We validate the intuition of segmentation being stronger than depth estimation for estimating object boundaries, even compared to depth from multi-view camera systems~\cite{yin2019hierarchical}, thus demonstrating the importance of leveraging this strength (Tab.~\ref{tab:morph}). 
We use the distance between segmentation and depth's edges as a measurement of their consistency. 
Since this measurement is not differentiable, we can not directly optimize it as a loss.\linebreak 
Rather, it is optimized as a ``greedy search", such that we iteratively construct a local optimum \textit{augmented} disparity map under the proposed measurement and penalize its discrepancy with the original prediction.
The construction of augmented depth map is done via a modified Beier–Neely morphing algorithm\cite{ucicr1992feature}. 
In this way, the estimated depth map gradually becomes more consistent with the segmentation edges within the scene, as demonstrated in Fig.~\ref{fig:teaser}. 

Since we use predicted semantics labels\cite{zhu2019improving}, noise is inevitably inherited. 
To combat this, we  develop several techniques to stabilize training as well as improve performance.
We also notice recent stereo-based self-supervised methods ubiquitously possess ``bleeding artifacts", which are fading borders around two sides of objects.\linebreak
We trace its cause to occlusions in stereo cameras near object boundaries and resolve by integrating a novel stereo occlusion mask into the loss, further enabling quality edges and subsequently facilitating our morphing technique.


Our contributions can be summarized as follows:

$\diamond$  We explicitly define and utilize the border constraint between semantic segmentation and depth estimation, resulting in depth more consistent with segmentation. 
    
$\diamond$  We alleviate the bleeding artifacts in prior depth  methods~\cite{godard2019digging, godard2017unsupervised, chen2019towards, pillai2019superdepth} via proposed stereo occlusion mask, furthering the depth quality near object boundaries. 
    
$\diamond$ We advance the state-of-the-art (SOTA) performance of the self-supervised monocular depth estimation task on the KITTI dataset, which for
    the first time matches SOTA supervised performance in the absolute relative  metric.

\Section{Related work}

\Paragraph{Self-supervised Depth Estimation}
Self-supervision has been a pivotal component in depth estimation~\cite{zhou2017unsupervised, vijayanarasimhan2017sfm, yin2018geonet}.  
Typically, such methods require only a monocular image in inference but are trained with video sequences, stereo images, or both. 
The key idea is to build pixel correspondences from a predicted depth map among images of different view angles then minimize a photometric reconstruction loss for all paired pixels. 
Video-based methods~\cite{zhou2017unsupervised, vijayanarasimhan2017sfm, yin2018geonet} require both depth map estimation and ego-motion. 
While stereo system~\cite{garg2016unsupervised, godard2017unsupervised} requires a pair of images captured simultaneously by cameras with known relative placement, reformulating depth estimation into disparity estimation. 

We note the photometric loss is subject to two general issues: $(1)$  When occlusions present, via stereo cameras or dynamic scenes in video, an incorrect pixel correspondence can be made yielding sub-optimal performance. $(2)$ There exists ambiguity in low-texture or color-saturated areas such as sky, road, tree leaves, and windows, thereby receiving a weak supervision signal. 
We aim to address $(1)$ by proposed stereo occlusion masking, and $(2)$ by leveraging additional explicit supervision from semantic segmentation.

\Paragraph{Occlusion Problem}
Prior works in video-based depth estimation~\cite{godard2019digging, vijayanarasimhan2017sfm, janai2018unsupervised, casser2019depth} have begun to address the occlusion problem. 
\cite{godard2019digging} suppresses occlusions by selecting pixels with a minimum photometric loss in consecutive frames. 
Other works \cite{vijayanarasimhan2017sfm, janai2018unsupervised} leverage optical flow to account for object and scene movement. 
In comparison, occlusion in stereo pairs has not received comparable attention in SOTA methods. 
Such occlusions often result in bleeding depth artifacts when (self-)supervised with photometric loss. 
\cite{godard2017unsupervised} partially relieves the bleeding artifacts via a left-right consistency term. Comparatively, \cite{pillai2019superdepth, yang2018deep} incorporates a regularization onto the depth magnitude to suppress the artifacts. 

In our work, we propose an efficient occlusion masking based only on a single estimated disparity map, which significantly improves estimation convergence and qualities around dynamic objects' border (Sec.~\ref{sec:occlusion_masking}). 
Another positive side effect is improved edge maps, which facilitates our proposed semantic-depth edge consistency (Sec.~\ref{sec:explicit_consistency_section}).

\Paragraph{Using Additional Modalities}
To address weak supervision in low-texture regions, prior work has begun incorporating modalities such as surface normal~\cite{yang2017unsupervised}, semantic segmentation~\cite{ramirez2018geometry, chen2019towards, wang2015towards, mousavian2016joint}, optical flow~\cite{vijayanarasimhan2017sfm, janai2018unsupervised} and stereo matching proxies~\cite{watson2019self, tosi2019learning}. 
For instance,~\cite{yang2017unsupervised} constrains the estimated depth to be more consistent with predicted surface normals. 
While~\cite{watson2019self, tosi2019learning} leverage proxy disparity labels produced by Semi-Global Matching (SGM) algorithms~\cite{hirschmuller2005accurate, hirschmuller2007stereo}, which serve as additional psuedo ground truth supervision. 
In our work, we provide a novel study focusing on constraints from the shared borders between segmentation and depth. 

\begin{figure*}
    \centering
     \figvspace
    \includegraphics[width = 0.95\linewidth]{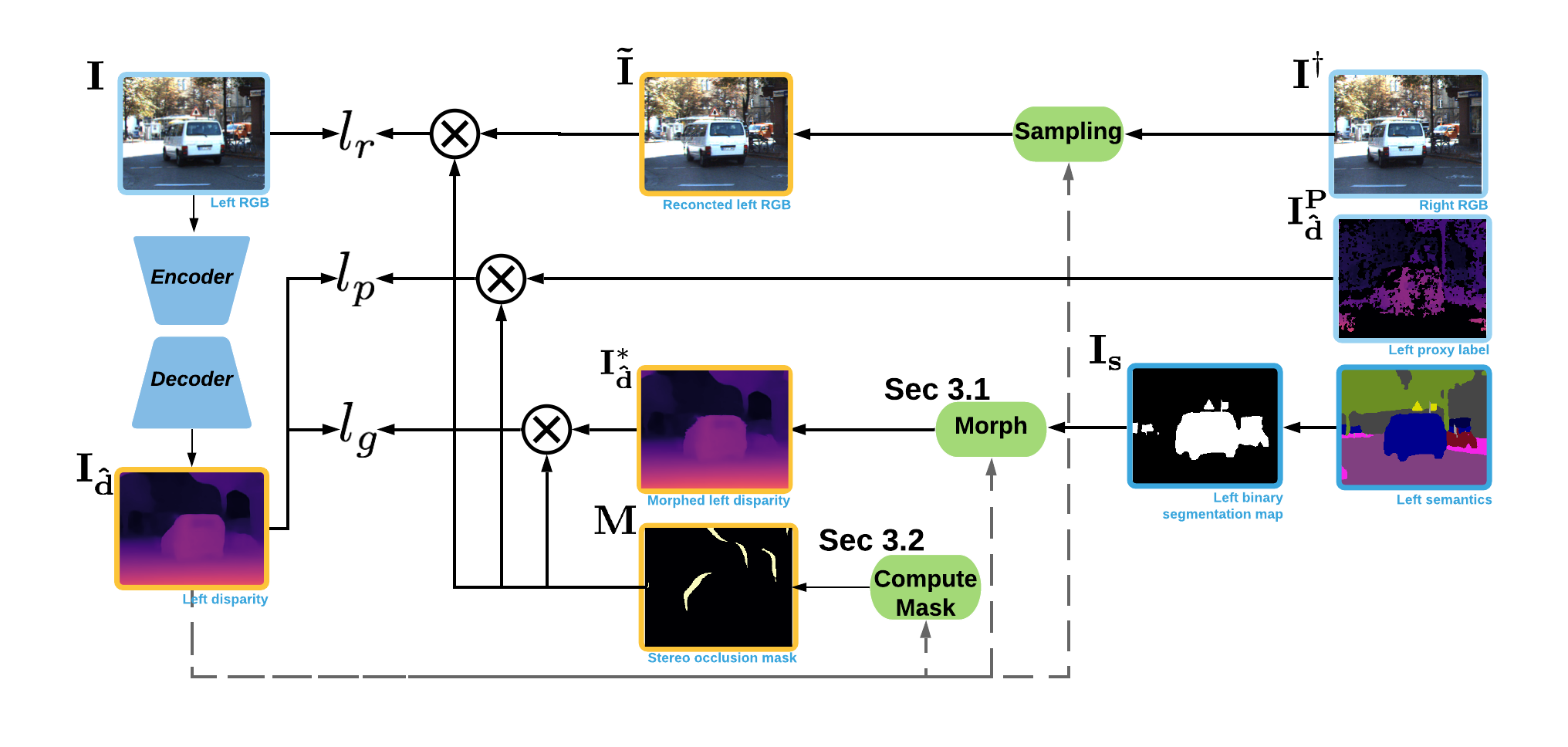}
    \vspace{-7mm}
    \caption{\textbf{Framework Overview}. The blue box indicates input while yellow box indicates the estimation. 
    The encoder-decoder takes only a left image $\mathbf{I}$, to predict the corresponding disparity $\mathbf{I_{\hat{d}}}$ which will be converted to depth map $\mathbf{I_d}$. The prediction is supervised via a photometric reconstruction loss $l_r$, morph loss $l_g$, and stereo matching proxy loss $l_p$.
    }
    \figvspace
    \label{fig:framework}
\end{figure*}

\Paragraph{Using Semantic Segmentation for Depth}
The relationship between depth and semantic segmentation is fundamentally different from the aforementioned modalities. Specifically, semantic segmentation does not inherently hold a definite mathematical relationship with depth. 
In contrast, surface normal can be interpreted as normalized depth gradient in 3D space;  
disparity possesses an inverse linear relationship with depth;
and optical flow can be decomposed into object movement, ego-motion, and depth estimation. 
Due to the vague relationship between semantic segmentation and depth, prior work primarily use it in an {\it implicit} manner.

We classify the uses of segmentation for depth estimation into three categories. 
Firstly, share weights between semantics and depth branches as in~\cite{chen2019towards, wang2015towards}.
Secondly, mix semantics and depth features as in~\cite{wang2015towards, mousavian2016joint, zhang2019pattern}. 
For instance,~\cite{wang2015towards, mousavian2016joint} use a conditional random field to pass information between modalities. 
Thirdly, \cite{kendall2018multi, ramirez2018geometry} opt to model the statistical relationship between segmentation and depth. 
\cite{kendall2018multi} specifically models the uncertainty of segmentation and depth to re-weight themselves in the loss function. 

Interestingly, no prior work has leveraged the border consistency naturally existed between segmentation and depth. 
We emphasize that leveraging this observation has two difficulties. 
First, segmentation and depth only share partial borders. 
Secondly, formulating a differentiable function to link binarized borders to continuous semantic and depth prediction remains a challenge.
Hence, designing novel approaches to address these challenges is our contribution to an explicit segmentation-depth constraint.



\Section{The Proposed Method}
\label{sec:method}
We observe recent self-supervised depth estimation methods\cite{watson2019self} preserve deteriorated object borders compared to semantic segmentation methods\cite{zhu2019improving} (Tab.~\ref{tab:morph}).
It motivates us to explicitly use segmentation borders as a constraint in addition to the typical photometric loss. \linebreak
We propose an edge-edge consistence loss $l_c$~(Sec.~\ref{sec:edgetoedge}) between depth map and segmentation map.
However, as the $l_c$ is not differentiable, we circumvent it by constructing an optimized depth map $\mathbf{I_d^{*}}$ and penalizing its difference with original prediction $\mathbf{I_d}$ (Sec.~\ref{sec:lossfunction}). 
This construction is accomplished via a novel morphing algorithm~(Sec.~\ref{sec:depthmorphing}). 
Additionally, we resolve bleeding artifacts~(Sec.~\ref{sec:occlusion_masking}) for improved border quality and rectify batch normalization layer statistics via a finetuning strategy (Sec.~\ref{sec:lossfunction}).
As in Fig.~\ref{fig:framework}, our method consumes stereo image pairs and precomputed semantic labels~\cite{zhu2019improving} in training, while only requiring a monocular RGB image at inference. It predicts a disparity map $\mathbf{I_{\hat{d}}}$ and then converted to depth map $\mathbf{I_{d}}$ given baseline $b$ and focal length $f$ under relationship $\mathbf{I_{d}} = \frac{f\cdot b}{\mathbf{I_{\hat{d}}}}$. 

\SubSection{Explicit Depth-Segmentation Consistency}
\label{sec:explicit_consistency_section}

To explicitly encourage estimated depth to agree with its segmentation counterpart on their edges, we propose two steps.
We first extract matching edges from segmentation $\mathbf{I_s}$ and corresponding depth map $\mathbf{I_d}$ (Sec.~\ref{sec:edgetoedge}).
Using these pairs, we propose a continuous morphing function to warp all depth values in its inner-bounds~(Sec.~\ref{sec:depthmorphing}), such that depth edges are aligned to semantic edges while preserving the continuous integrity of the depth map.

\SubSubSection{Edge-Edge Consistency}
\label{sec:edgetoedge}
In order to define the edge-edge consistency, we must firstly extract the edges from both the segmentation map $\mathbf{I_s}$ and depth map $\mathbf{I_d}$. 
We define $\mathbf{I_s}$ as a binary foreground-background segmentation map, whereas the depth map $\mathbf{I_d}$ consists of continuous depth values.
Let us denote an edge $\mathbf{T}$ as the set of pixel $\mathbf{p}$ locations such that:
\begin{equation}
\mathbf{T} = \Big\{\mathbf{p}\, \mid\,\norm{\frac{\partial \mathbf{I}(\mathbf{p})}{\partial \mathbf{x}}} > k_1\Big\},
\end{equation}
where $\frac{\partial \mathbf{I}(\mathbf{p})}{\partial \mathbf{x}}$ is a 2D image gradient at $\mathbf{p}$ and $k_1$ is a hyperparameter controlling necessary gradient intensity to constitute an edge.
In order to highlight clear borders in close-range objects, the depth edge $\mathbf{T_d}$ is extracted from the disparity map $\mathbf{I_{\hat{d}}}$ instead of $\mathbf{I_d}$.
Given an arbitrary segmentation edge point $\mathbf{q} \in \mathbf{T_s}$, we denote $\delta(\mathbf{q}, \mathbf{T_d})$ as the distance between $\mathbf{q}$ to its closest point in depth edge $\mathbf{T_d}$:
\begin{equation}
     \delta(\mathbf{q}, \mathbf{T_d}) = \min_{\left\{\mathbf{p} \mid \mathbf{p} \in \mathbf{T_d} \right\}} \norm{\mathbf{p} - \mathbf{q}}.
\end{equation}
Since the correspondence between segmentation and depth edges do not strictly follow an one-one mapping, we limit it to a predefined local range.
We denote the valid set $\mathbf{\Gamma}$ of segmentation edge points $\mathbf{q} \in \mathbf{T_s}$ such that: 
\begin{equation}
    \mathbf{\Gamma}(\mathbf{T_s} \mid \mathbf{T_d}) = \left\{ \mathbf{q} \mid \forall \mathbf{q} \in \mathbf{T_s},\,\,  \delta(\mathbf{q}, \mathbf{T_d}) < k_2\right\},
\end{equation}
where $k_2$ is a hyperparamter controlling the maximum distance allowed for association. 
For notation simplicity, we denote $\mathbf{\Gamma^d_s} = \mathbf{\Gamma}(\mathbf{T_s} \mid \mathbf{T_d})$.
Then the consistency $l_c$ between the segmentation $\mathbf{T_s}$ and depth $\mathbf{T_d}$ edges is as: 
\begin{equation}
    l_c(\mathbf{\Gamma}(\mathbf{T_s} \mid \mathbf{T_d}),~ \mathbf{T_d}) = \frac{1}{\norm{ \mathbf{\Gamma_s^d} }} \sum_{\mathbf{q}\in \mathbf{\Gamma_s^d}} \delta(\mathbf{q}, \mathbf{T_d}).
    \label{eq:consDef}
\end{equation}
Due to the discretization used in extracting edges from $\mathbf{I_s}$ and $\mathbf{I_d}$, it is difficult to directly optimize $l_c(\mathbf{\Gamma_s^d},~\mathbf{T_d})$.
Thus, we propose a continuous morph function ($\phi$ and $g$ in Sec.~\ref{sec:depthmorphing}) to produce an augmented depth $\mathbf{I^*_d}$, with a corresponding depth edge $\mathbf{T^*_d}$ that minimizes:
\begin{equation}
\begin{aligned}
l_c(\mathbf{\Gamma}(\mathbf{T_s} \mid \mathbf{T_d}),~ \mathbf{T^*_d}).
\end{aligned}
\label{eq:local_min}
\end{equation}
Note that the $l_c$ loss is asymmetric. 
Since the segmentation edge is more  reliable, we prefer to use $l_c(\mathbf{\Gamma_s^d},~\mathbf{T^*_d})$ rather than its inverse mapping direction of $l_c(\mathbf{\Gamma_d^s},~\mathbf{T^*_s})$.

\SubSubSection{Depth Morphing}
\label{sec:depthmorphing}

\begin{figure}
    \centering
    \includegraphics[width = \linewidth]{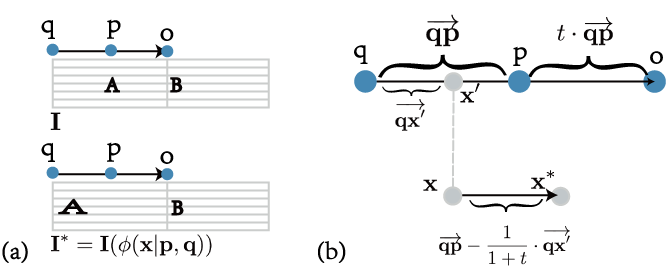}
    \vspace{-5mm}
    \caption{
        The \textbf{morph function} $\phi(\cdot)$  morphs a pixel $\mathbf{x}$ to pixel $\mathbf{x^*}$, via Eq.~\ref{eq:point_sampling} and~\ref{eq:point_projection}.
        (a) A source image $\mathbf{I}$ is morphed to $\mathbf{I^*}$ by applying $\phi(\mathbf{x}| \mathbf{q}, \mathbf{p})$ to every pixel $\mathbf{x}\in \mathbf{I}^*$ with the closest pair of segmentation $\mathbf{q}$ and depth $\mathbf{p}$ edge points. 
        (b) we show each term's geometric relationship.
        The morph warps $\mathbf{x}$ around $\protect\overrightarrow{\mathbf{qo}}$ to $\mathbf{x^*}$ around $\protect\overrightarrow{\mathbf{po}}$. 
       Point $\mathbf{o}$ is controlled by term $t$ in the extended line of $\protect\overrightarrow{\mathbf{qp}}$.
        }
    \label{fig:alpha}
    \figvspace
\end{figure}

In the definition of consistence measurement $l_c$ in  Eq.~\eqref{eq:local_min}, we acquire a set of associations between segmentation and depth border points. 
We denote this set as $\mathbf{\Omega}$:
\begin{equation}
    \mathbf{\Omega} = \Big\{ \mathbf{p} \mid \argmin_{\{\mathbf{p} \mid \mathbf{p} \in \mathbf{T_d}\}} \norm{\mathbf{p} - \mathbf{q}}, \mathbf{q} \in \mathbf{\Gamma_s^d} \Big\}.
\end{equation}
Associations in $\mathbf{\Omega}$ imply depth edge $\mathbf{p}$ should be adjusted towards segmentation edge $\mathbf{q}$ to minimize consistence measurement $l_c$. This motivates us to design a local morph function $\phi(\cdot)$ which maps an arbitrary point $\mathbf{x}$ near a segmentation point $\mathbf{q}\in\mathbf{\Gamma_s^d}$ and associated depth point $\mathbf{p}\in\mathbf{\Omega}$ to:
\begin{equation}
    \mathbf{x^*} = \phi(\mathbf{x} \mid \mathbf{q}, \mathbf{p}) =  \mathbf{x} +  \protect\overrightarrow{\mathbf{qp}} - \frac{1}{1+t}\cdot \protect\overrightarrow{\mathbf{qx'}} ,
    \label{eq:point_sampling}
\end{equation}
where hyperparameter $t$ controls sample space illustrated in Fig.~\ref{fig:alpha}, and $\mathbf{x'}$ denotes the point projection of $\mathbf{x}$ onto  $\protect\overrightarrow{\mathbf{qp}}$:
\begin{equation}
    \mathbf{x'} = \mathbf{q} + (\protect\overrightarrow{\mathbf{qx}} \cdot \hat{\mathbf{qp}})\cdot \hat{\mathbf{qp}},
    \label{eq:point_projection}
\end{equation}
where $\hat{\mathbf{qp}}$ is the unit vector of the associated edge points.
We illustrate a detailed example of  $\phi(\cdot)$ in Fig.~\ref{fig:alpha}.

To promote smooth and continuous morphing, we further define a more robust morph function $g(\cdot)$, applied to every pixel $\mathbf{x}\in \mathbf{I^*_d}$ as a distance-weighted summation of all morphs $\phi(\cdot)$ for each associated pair $(\mathbf{q}, \mathbf{p}) \in (\mathbf{\Gamma_s^d}, \mathbf{\Omega})$:
\begin{align}  
\resizebox{0.9\hsize}{!}{
     $g(\mathbf{x} \mid \mathbf{q}, \mathbf{p}) = \sum_{i = 0}^{i = |\mathbf{\Omega}|} \frac{w(d_i)}{\sum_{j = 0}^{j = |\mathbf{\Omega}|}w (d_j)} \cdot h(d_i) \cdot \phi(\mathbf{x} \mid \mathbf{p}_i, \mathbf{q}_i)$,
     }
\end{align}
where $d_i$ is the distance between $\mathbf{x}_i$ and edge segments $\protect\overrightarrow{\mathbf{q}_i \mathbf{p}_i}$. $h(\cdot)$ and $w(\cdot)$ are distance-based weighting functions: 
$w(d_i) = (\frac{1}{m_3 + d_i})^{m_4}$, and $ h(d_i) = \text{Sigmoid}(-m_1 \cdot (d_i - m_2))$, 
where $m_1, m_2, m_3$, $m_4$ are predefined hyperparameters. $w(\cdot)$ is a relative weight compromising morphing among multiple pairs, while $h(\cdot)$ acts as an absolute weight ensuring each pair only affects local area. 
Implementation wise, $h(\cdot)$ makes pairs beyond $\Sim7$ pixels negligible, facilitating $g(\mathbf{x} \mid \mathbf{q},\mathbf{p})$ linear computational complexity.

In summary, $g(\mathbf{x} \mid \mathbf{q},\mathbf{p})$ can be viewed as a more general Beier–Neely~\cite{ucicr1992feature} morph, due to inclusion of $h(\cdot)$.
We align depth map better to segmentation via applying $g(\cdot)$ morph to pixels of its disparity map $\mathbf{x}\in \mathbf{I^*_{\hat{d}}}$, creating a segmentation-augmented disparity map $\mathbf{I_{\hat{d}}^*}$:
\begin{equation}
    \begin{aligned}
        &\mathbf{I_{\hat{d}}^*}(\mathbf{x}) = \mathbf{I_{\hat{d}}}(g(\mathbf{x} \mid \mathbf{q}, \mathbf{p})) \\
        \vdash \quad &\forall (\mathbf{p}, \mathbf{q}) \, \in (\mathbf{\Omega}, \Gamma), \,\,\mathbf{p} = \phi(\mathbf{q}).
    \end{aligned}
    \label{eq:i_d_hat}
\end{equation}
Next we may transform the edge-to-edge consistency term $l_{c}$ into the minimization of difference between $\mathbf{I_{\hat{d}}}$ and the segmentation-augmented $\mathbf{I_{\hat{d}}^*}$, as detailed in Sec.~\ref{sec:lossfunction}. 
A concise proof of $\mathbf{I_d^*}$ as local minimum of $l_{c}$ under certain condition is in the supplementary material (\textbf{Suppl.}).


\SubSection{Stereo Occlusion Mask}
\label{sec:occlusion_masking}

\begin{figure}
  \centering
  \begin{tabular}[b]{c c c}
    \includegraphics[width=.25\linewidth]{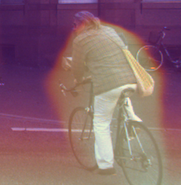} & 
    \includegraphics[width=.25\linewidth]{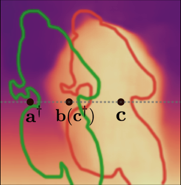} & \includegraphics[width=.25\linewidth]{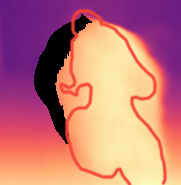}  \vspace{-1mm}\\
    \small(a) & \small(b) & \small(c)
  \end{tabular}
  \vspace{-2mm}
  \caption{
  (a) Overlays disparity estimation over the input image showing typical \textbf{bleeding artifacts}. 
  (b) We denote the \textbf{\textcolor{darkred}{red}} object contour from the left  view $\mathbf{I}$ and \textbf{\textcolor{darkgreen}{green}} object contour from the right view $\mathbf{I}^\dagger$. 
  Background point $\mathbf{b}$ is visible in the left view, yet its corresponding right point $\mathbf{b}^\dagger$ is occluded by an object point $\mathbf{{c}^\dagger}$. 
  Thus, this point is incorrectly supervised by photometric loss $l_r$ to look for the nearest background pixel (\emph{e.g.}, $\mathbf{a}^\dagger$) leading to a bleeding artifact in (a). 
  (c) We depict occluded region detected via Eq.~\ref{eq:stereo_occlusion}.
  }
  \label{fig:stereo_mask}
  \figvspace
\end{figure}

Bleeding artifacts are a common difficulty in self-supervised stereo methods~\cite{godard2019digging, godard2017unsupervised, chen2019towards, pillai2019superdepth}. 
Specifically, bleeding artifacts refer to instances where the estimated depth on surrounding foreground objects wrongly expands outward to the background region.
However, few works provide detailed analysis of its cause. 
We illustrate the effect and an overview of our stereo occlusion mask in Fig.~\ref{fig:stereo_mask}.

Let us define a point $\mathbf{b} \in \mathbf{I_d}$ near the boundary of an object and corresponding point $\mathbf{b}^\dagger \in \mathbf{I^\dagger_d}$ in the right stereo view.
When point $\mathbf{{b}^{\dagger}}$ is occluded by a foreground point $\mathbf{{c}^\dagger}$ in the right stereo, a photometric loss will seek a similar non-occluded point in the right stereo, \emph{e.g.}, the objects' left boundary $\mathbf{{a}^{\dagger}}$, since no exact solution may exist for occluded pixels. 
Therefore, the disparity value at point $\mathbf{\text{b}}$ will be $\hat{d}_{\mathbf{{b}}}^* = \norm{\protect\overrightarrow{\mathbf{{a}^{\dagger}\text{b}}}} = x_{\mathbf{\text{b}}} - x_{\mathbf{{a}^{\dagger}}}$, where $x$ is the horizontal location. \linebreak
Since background is assumed farther away than foreground points, generally a false supervision has the quality such that the occluded background disparity will be significantly larger than its (unknown) ground truth value.
As $\mathbf{b}$ approaches $\mathbf{a}^\dagger$ the effect is lessened, creating a fading effect. 

To alleviate the bleeding artifacts, we form an occlusion indicator matrix $\mathbf{M}$ such that $\mathbf{M}(x, y) = 1$ if the pixel location $(x, y)$ has possible occlusions in the stereo view. \linebreak
For instance, in the left stereo image $\mathbf{M}$ is defined as:
\begin{align}
\resizebox{0.88\hsize}{!}{
   $ \mathbf{M}(x, y) = 
    \begin{cases}
    1 &\underset{i \in \left(0, W - x \right]}{\min} \left(\mathbf{I_{\hat{d}}}(x + i, y) - \mathbf{I_{\hat{d}}}(x, y)  -i  \right) \geq k_3\\
    0              & \text{otherwise},
    \end{cases}
    $
    }
    \label{eq:stereo_occlusion}
\end{align}
where $W$ denotes predefined search width and $k_3$ is a threshold controlling thickness of the mask.
The disparity value in the left image represents the horizontal left distance of each pixel to be moved. 
As the occlusion is due to pixels in its right, we intuitively perform our search in one direction.
Additionally, we can view occlusion as when neighbouring pixels on its right move too much left and cover itself. 
In this way, occlusion can be detected as $\underset{i \in \left(0, W - x \right]}{\min} \left(\mathbf{I_{\hat{d}}}(x + i, y) - \mathbf{I_{\hat{d}}}(x, y)  -i  \right) \geq 0$. 
Considering bleeding artifacts in Fig.~\ref{fig:stereo_mask}, we use $k_3$ to counter large incorrect disparity values of occluded background pixels. 
The regions indicated by $\mathbf{M}$ are then masked when computing a reconstruction loss (Sec.~\ref{sec:lossfunction}).

\SubSection{Network and Loss Functions}
Our network is comprised of an encoder-decoder, identical to the baseline~\cite{watson2019self}. It takes in a monocular RGB image and predicts corresponding disparity map which is later converted to depth map under known camera parameters.
\SubSubSection{Loss Functions}
\label{sec:lossfunction}

The overall loss function is comprised of three terms:
\begin{equation}
    l = l_r(\mathbf{I_{\hat{d}}}(\mathbf{x})) + \lambda_2 l_g(\mathbf{I_{\hat{d}}}(\mathbf{x})) + \lambda_1 l_p(\mathbf{I_{\hat{d}}}(\mathbf{x})),
    \label{eq:jointloss}
\end{equation}
where $l_r$ denotes a photometric reconstruction loss, $l_g$ a morphing loss, $l_p$ a stereo proxy loss~\cite{watson2019self}, and $\mathbf{x}$ are the non-occluded pixel locations, {\it i.e.}, $\{\mathbf{x} \mid \mathbf{M}(\mathbf{x}) = 0\}$. 
$\lambda_1$ and $\lambda_2$ are the weights of terms. 
We emphasize that exclusion will not prevent learning of object borders. 
\textit{E.g.,} in Fig.~\ref{fig:stereo_mask}(c), although the pixel $\mathbf{b}$ in cyclist's left border is occluded, the network can still learn to estimate depth from a visible and highly similar pixel $\mathbf{a}^{\dagger}$ in the stereo counterpart, as both left and right view images are respectively fed into the encoder in training, similar to prior self-supervised works~\cite{watson2019self, godard2019digging}. 

Following~\cite{godard2019digging}, we define the $l_r$ reconstruction loss as:
\begin{equation}
    \resizebox{\hsize}{!}{
    $l_{r}\left(\mathbf{I_{\hat{d}}}(\mathbf{x})\right)=\alpha \frac{1-\operatorname{SSIM}\left(\mathbf{I}(\mathbf{x}), \tilde{ \mathbf{I}}(\mathbf{x})\right)}{2}+(1-\alpha) | \mathbf{I}(\mathbf{x})-\tilde{\mathbf{I}}(\mathbf{x})|,$}
    \label{eq:photometricl}
\end{equation}
which consists of a pixel-wise mix of SSIM~\cite{wang2004image} and $L_1$ loss between an input left image $\mathbf{I}$ versus the reconstructed left image $\mathbf{\tilde{I}}$, which is re-sampled according to predicted disparity $\mathbf{I_{\hat{d}}}$.
The $\alpha$ is a weighting  hyperparameter as in~\cite{godard2017unsupervised, watson2019self}.

We minimize the distance between depth and segmentation edges by steering the disparity $\mathbf{I_{\hat{d}}}$ to approach the semantic-augmented disparity $\mathbf{I_{\hat{d}}^*}$ (Eq.~\ref{eq:i_d_hat}) in a logistic loss:
\begin{equation}
        l_g(\mathbf{I_{\hat{d}}}(\mathbf{x})) = \mathbf{w(\mathbf{I_{\hat{d}}(x)})} \cdot \log(1 + |\mathbf{I_{\hat{d}}^*}(\mathbf{x}) - \mathbf{I_{\hat{d}}}(\mathbf{x})|),
        \label{eq:greedyloss}
\end{equation}
where $\mathbf{w}(\cdot)$ is a  function to downweight image regions with low variance. 
It is observed that the magnitude of the photometric loss (Eq.~\ref{eq:photometricl}) varies significantly between textureless and rich texture image regions, whereas the morph loss (Eq.~\ref{eq:greedyloss}) is primarily dominated by the border consistency.
Moreover, the morph is itself dependent on an \textit{estimated} semantic psuedo ground truth $\mathbf{I_s}$~\cite{zhu2019improving} which may include noise. 
In consequence, we only apply the loss when the photometric loss is comparatively improved. 
Hence, we define the weighting function $\mathbf{w}(\cdot)$ as:
\begin{equation}
    \resizebox{0.88\hsize}{!}{$
    \mathbf{w}(\mathbf{I_{\hat{d}}}(\mathbf{x})) = \left\{
    \begin{array}{ll}
    \text{Var}(\mathbf{I})(\mathbf{x}) &\text{If} \hspace{0.5em} l_r(\mathbf{I^*_{\hat{d}}}(\mathbf{x})) < l_r(\mathbf{I_{\hat{d}}}(\mathbf{x})) \\
    0 \quad &\text{otherwise,}
    \end{array}\right.$
    }
    \label{eq:stabilizing_w}
\end{equation}
where $\text{Var}(\mathbf{I})$ computes pixel-wise RGB image variance in a $3\times3$ local window.
Note that when a noisy semantic estimation $\mathbf{I_s}$ causes $l_r$ to degrade, the pixel location is ignored. 

Following \cite{watson2019self}, we incorporate a stereo proxy loss $l_p$ which we find helpful in neutralizing noise in estimated semantics labels, defined similarly to Eq.~\ref{eq:greedyloss} as:
\begin{equation}
    \resizebox{\hsize}{!}{$
    l_p(\mathbf{I_{\hat{d}}}(\mathbf{x})) = \left\{
    \begin{array}{ll}
    \text{log}(1 + |\mathbf{I_{\hat{d}}^p} - \mathbf{I_{\hat{d}}}|) &\text{If} \hspace{0.5em} l_r(\mathbf{I_{\hat{d}}^p} (\mathbf{x})) < l_r(\mathbf{I_{\hat{d}}}(\mathbf{x})) \\
    0 \quad &\text{otherwise,}
    \end{array}\right.$
    }
    \label{eq:proxyloss}
\end{equation}
where $\mathbf{I_{\hat{d}}^p}$ denotes the stereo matched proxy label generated by the Semi-Global Matching (SGM)~\cite{hirschmuller2005accurate, hirschmuller2007stereo} technique.

\begin{table*}
    \centering
    \resizebox{\linewidth}{!}{%
    \begin{tabular}{|c|c|c|c|c|c|c|c|c|c|c|c|c|}
        \hline
         Cita. & Method  & PP & Data &H $\times$ W & Size (Mb) &\cellcolor[RGB]{222, 164, 151}Abs Rel &\cellcolor[RGB]{222, 164, 151} Sq Rel &\cellcolor[RGB]{222, 164, 151} RMSE &\cellcolor[RGB]{222, 164, 151} RMSE log & \cellcolor[RGB]{155, 187, 228}$\delta  < 1.25 $ & \cellcolor[RGB]{155, 187, 228}$\delta  < 1.25^2 $ & \cellcolor[RGB]{155, 187, 228}$\delta  < 1.25^3 $\\
         \hline
         \cite{yang2018deep} & Yang \textit{et al}. & $\color[RGB]{66, 146, 41}\checkmark$& $\text{D}^{\dagger}\text{S}$ & $256 \times 512$ & -& $0.097$ & $0.734$ & $4.442$ & $0.187$ & $0.888$ & $0.958$ & $0.980$ \\
         \cite{guo2018learning} & Guo \textit{et al}. &  &$\text{D}^* \text{DS}$ & $256 \times 512$& $79.5$ &$0.097$ &$0.653$ &$4.170$& $0.170$& $0.889$& $\mathbf{0.967}$& $\mathbf{0.986}$ \\
         \cite{luo2018single} & Luo \textit{et al}.& &$\text{D}^* \text{DS}$ & $192 \times 640\,\, \text{crop}$& $1,562$ &$0.094$& $0.626$& $4.252$& $0.177$& $0.891$& $0.965$& $0.984$ \\
         \cite{kuznietsov2017semi} & Kuznietsov \textit{et al}. & &DS & $187 \times621$& $324.8$ & $0.113$ & $0.741$ & $4.621$ & $0.189$ & $0.862$ & $0.960$ & $\mathbf{0.986}$ \\
         \cite{fu2018deep} & Fu \textit{et al}. && D & $385 \times 513\,\, \text{crop}$& $399.7$ & $0.099$& $0.593$& $\mathbf{3.714}$& $\mathbf{0.161}$& $0.897$& $0.966$& $\mathbf{0.986}$\\
         \cite{lee2019big} & Lee \textit{et al}. && D & $352 \times 1,216$&$563.4$ & $\mathbf{0.091}$ & $\mathbf{0.555}$& $4.033$& $0.174$& $\mathbf{0.904}$& $\mathbf{0.967}$& $0.984$ \\
         \hline
         \hline
         \cite{godard2017unsupervised} & Godard \textit{et al}. & $\color[RGB]{66, 146, 41}\checkmark$ & S & $256 \times 512$&$382.5$ &$0.138$ &$1.186$& $5.650$& $0.234$& $0.813$ &$0.930$ &$0.969$ \\
         \cite{mehta2018structured} & Mehta \textit{et al}. & &S& $256 \times 512$&- &$0.128$ &$1.019$& $5.403$& $0.227$& $0.827$& $0.935$& $0.971$ \\
         \cite{poggi2018learning} & Poggi \textit{et al}. & $\color[RGB]{66, 146, 41}\checkmark$ & S & $256 \times 512$&$954.3$ &$0.126$ &$0.961$& $5.205$& $0.220$& $0.835$& $0.941$& $0.974$ \\
         \cite{zhan2018unsupervised} & Zhan \textit{et al}. &$\color[RGB]{184, 37, 25}\text{\ding{55}}$ & MS & $160 \times 608$&- &$0.135$& $1.132$& $5.585$& $0.229$& $0.820$ &$0.933$ &$0.971$ \\
         \cite{luo2018every} & Luo \textit{et al}. & & MS &$256 \times 832$&$160$ & $0.128$ & $0.935$ & $5.011$ & $0.209$ & $0.831$ & $0.945$ & $0.979$ \\
         \cite{pillai2019superdepth} & Pillai \textit{et al}. & $\color[RGB]{66, 146, 41}\checkmark$ & S & $384 \times 1,024$&- &$0.112$ &$0.875$& $4.958$& $0.207$& $0.852$& $0.947$& $0.977$ \\
         \cite{tosi2019learning} & Tosi \textit{et al}. & $\color[RGB]{66, 146, 41}\checkmark$ & S & $256 \times 512\,\, \text{crop}$&$511.0$ & $0.111$ & $0.867$ & $4.714$ & $0.199$ & $0.864$ & $0.954$ & $0.979$ \\
         \cite{chen2019towards} & Chen \textit{et al}.& $\color[RGB]{66, 146, 41}\checkmark$& SC&$256 \times 512$&- &$0.118$ &$0.905$& $5.096$& $0.211$& $0.839$& $0.945$& $0.977$\\
         \cite{godard2019digging} & Godard \textit{et al}.& $\color[RGB]{66, 146, 41}\checkmark$ & MS & $320 \times 1,024$&$59.4$ &$0.104$ &$0.775$& $4.562$& $0.191$& $0.878$& $0.959$& $0.981$ \\
         \cite{watson2019self} & Watson \textit{et al}. (ResNet18)& $\color[RGB]{66, 146, 41}\checkmark$ & S &  $320 \times 1,024$&$59.4$ & $0.099$ &$0.723$& $4.445$& $0.187$& $0.886$& $0.962$& $0.981$ \\
          &Ours (ResNet18)& $\color[RGB]{66, 146, 41}\checkmark$ & $\text{S}\text{C}^{\dagger}$ &  $320 \times 1,024$&$59.4$ &   $0.097$  &   $0.675$  &   $4.350$  &   $0.180$  &   $0.890$  &   $0.964$  &   $0.983$ \\
         \cite{watson2019self} & Watson \textit{et al}. (ResNet50)& $\color[RGB]{66, 146, 41}\checkmark$ & S &  $320 \times 1,024$&$138.6$ & $0.096$ & $0.710$ & $4.393$ & $0.185$ & $0.890$ & $0.962$ & $0.981$ \\
         & Ours (ResNet50)& $\color[RGB]{66, 146, 41}\checkmark$ & $\text{S}\text{C}^{\dagger}$  &  $320 \times 1,024$&$138.6$& $\mathbf{0.091}$  &   $\mathbf{0.646}$  &   $\mathbf{4.244}$  &   $\mathbf{0.177}$  &   $\mathbf{0.898}$  &   $\mathbf{0.966}$  &   $\mathbf{0.983}$\\
         \hline
    \end{tabular}
    }
    \vspace{-2mm}
    \caption{\textbf{Depth Estimation Performance}, on KITTI Stereo $2015$ dataset~\cite{geigerwe} eigen splits~\cite{eigen2014depth} capped at $80$ meters. The Data column denotes: $\text{D}$ for ground truth depth,  $\text{D}^{\dagger}$ for SLAM auxiliary data, $\text{D}^*$ for synthetic depth labels, S for stereo pairs, M for monocular video, C for segmentation labels, C$^{\dagger}$ for predicted segmentation labels. PP denotes post-processing. Size refers to the model size in Mb, which could be different depend on implementation language.}
    \label{tab:sota}
    \figvspace
\end{table*}

\Paragraph{Finetuning Loss:}
We further finetune the model to regularize the batch normalization~\cite{ioffe2015batch} statistics to be more consistent to an identity transformation.
As such, the prediction becomes less sensitive to the exponential moving average, following inspiration from~\cite{Singh_2019_ICCV} denoted as:
    $l_{\text{bn}} = \norm{\mathbf{I_{\hat{d}}}(\mathbf{x}) - \mathbf{I_{\hat{d}}^{'}}(\mathbf{x})}^2$,
where $\mathbf{I_{\hat{d}}}$ and $\mathbf{I_{\hat{d}}^{'}}$ denote predicted disparity with and without batch normalization, respectively.

\SubSubSection{Implementation Details} 
\label{sec:implementation}
We use PyTorch~\cite{paszke2017automatic} for training, and preprocessing techniques of~\cite{godard2019digging}.
To produce the stereo proxy labels, We follow \cite{watson2019self}.
Semantic segmentation is precomputed via~\cite{zhu2019improving}, in an ensemble way with default settings at a resolution of $320\times 1{,}024$. 
Using semantics definition in Cityscapes~\cite{cordts2016cityscapes}, we set object, vehicle, and human categories as foreground, and the rest as background. 
This allows us to convert a semantic segmentation mask to a binary segmentation mask  $\mathbf{I_s}$.
We use a learning rate of $1e^{-4}$ and train the joint loss (Eq.~\ref{eq:jointloss}) for $20$ epochs, starting with ImageNet~\cite{deng2009imagenet} pretrained weights. 
After convergence, we apply $l_{\text{bn}}$ loss for $3$ epochs at a learning rate of $1e^{-5}$.\linebreak
We set $t=\lambda_1=1$, $\lambda_2=5$, $k_1=0.11$, $k_2=20$, $k_3 = 0.05$, $m_1 = 17$, $m_2 = 0.7$, $m_3 = 1.6$, $m_4=1.9$, and $\alpha = 0.85$. Our source code is hosted at \url{http://cvlab.cse.msu.edu/project-edgedepth.html}.

\Section{Experiments}

We first present the comprehensive comparison on the KITTI benchmark, then analyze our results, and finally ablate various design choices of the proposed method.

\Paragraph{KITTI Dataset:}
We compare our method against SOTA works on KITTI Stereo 2015 dataset~\cite{geigerwe}, a comprehensive urban autonomous driving dataset providing stereo images with aligned LiDAR data. 
We utilize the eigen splits, evaluated with the standard seven KITTI metrics~\cite{eigen2014depth} with the crop of Garg~\cite{garg2016unsupervised} and a standard distance cap of $80$ meters~\cite{godard2017unsupervised}. 
Readers can refer to~\cite{eigen2014depth,geigerwe} for explanation of used metrics.

\Paragraph{Depth Estimation Performance:} 
We show a comprehensive comparison of our method to the SOTA in Tab.~\ref{tab:sota}. 
Our framework outperforms prior methods on each of the seven metrics. 
For a fair comparison, we utilize the same network structure as~\cite{godard2019digging, watson2019self}. 
We consider that approaching the performance of supervised methods is an important goal of self-supervised techniques.
Notably, our method is {\it the first self-supervised method matching SOTA supervised performance}, as seen in the absolute relative metric in Tab.~\ref{tab:sota}.
Additionally, We emphasize our method improves on the $\delta < 1.25$ from $0.890$ to $0.898$, thereby reducing the gap between supervised and unsupervised methods by relative $\Sim60\%$ $(=1 - \frac{0.904 - 0.898}{0.904 - 0.890})$. \linebreak
We further demonstrate a consistent performance gain with two variants of ResNet (Tab.~\ref{tab:sota}), demonstrating our method's robustness to the backbone architecture capacity. 

We emphasize our contributions are orthogonal to most methods including stereo and monocular training. 
For instance, we use noisy segmentation \textit{predictions}, which can be further enhanced by pairing with stronger segmentation or via segmentation annotations. 
Moreover, recall that we do not use the monocular training strategy of~\cite{godard2019digging} or additional stereo data such as Cityscapes, and utilize a substantially smaller network (\emph{e.g.}, $138.6$ vs.~$563.4$ MB~\cite{lee2019big}), thereby leaving more room for future enhancements.

\begin{table}[t!]
    \centering
    \small
    \setlength\tabcolsep{1pt}
    \resizebox{0.95\linewidth}{!}{%
    \begin{tabular}{|c|c|c|c|c|c|c|}
        \hline
         Method & Area &\cellcolor[RGB]{222, 164, 151}Abs Rel &\cellcolor[RGB]{222, 164, 151} Sq Rel &\cellcolor[RGB]{222, 164, 151} RMSE &\cellcolor[RGB]{222, 164, 151} RMSE log & \cellcolor[RGB]{155, 187, 228}$\delta  < 1.25 $\\
        \hline
         \multirow{3}{*}{Watson \textit{et al}. \cite{watson2019self}} 
        &$\text{O}$ &   $0.085$  &   $0.507$  &   $3.684$  &   $0.159$  &   $0.909$ \\
        &$\text{W}$ &   $0.096$  &   $0.712$  &   $4.403$  &   $0.185$  &   $0.890$ \\
        &$\text{N}$ &   $0.202$  &   $2.819$  &   $8.980$  &   $0.342$  &   $0.702$ \\
        \hline
         \multirow{3}{*}{Ours (ResNet50)} 
        &$\text{O}$ &   $0.081$  &   $0.466$  &   $3.553$  &   $0.152$  &   $0.916$  \\
        &$\text{W}$ &   $0.091$  &   $0.646$  &   $4.244$  &   $0.177$  &   $0.898$  \\
        &$\text{N}$ &   $0.192$  &   $2.526$  &   $8.679$  &   $0.324$  &   $0.712$  \\
        \hline
    \end{tabular}
    }
    \vspace{-2mm}
    \caption{\small{\textbf{Edge vs.~Off-edge Performance}. We evaluate the depth performance for {O-}off edge, {W-}whole image, {N-}near edge.}}
    \label{tab:edge_performance}
    \figvspace
\end{table}

\begin{figure}[t!]
    \centering
    \includegraphics[width = 0.95 \linewidth]{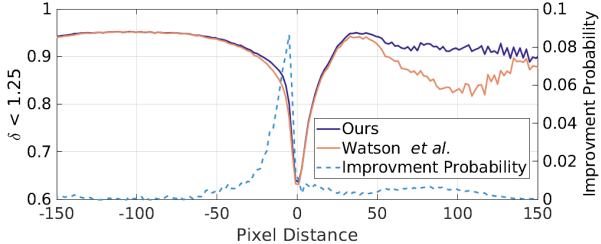}
    \vspace{-2mm}
    \caption{Left axis: Metric $\delta < 1.25$ as a function of distance off segmentation edges in background ($-x$) and foreground ($+x$), compared to~\cite{watson2019self}. Right axis: improvement distribution against distance.
    Our gain mainly comes from  near-edge background area but not restricted to it.
    } 
    \label{fig:delta_funciton_of_pixel_distnace}
    \figvspace
    \vspace{2mm}
\end{figure}

\begin{figure}[t!]
 \centering
   \includegraphics[width=0.95\linewidth]{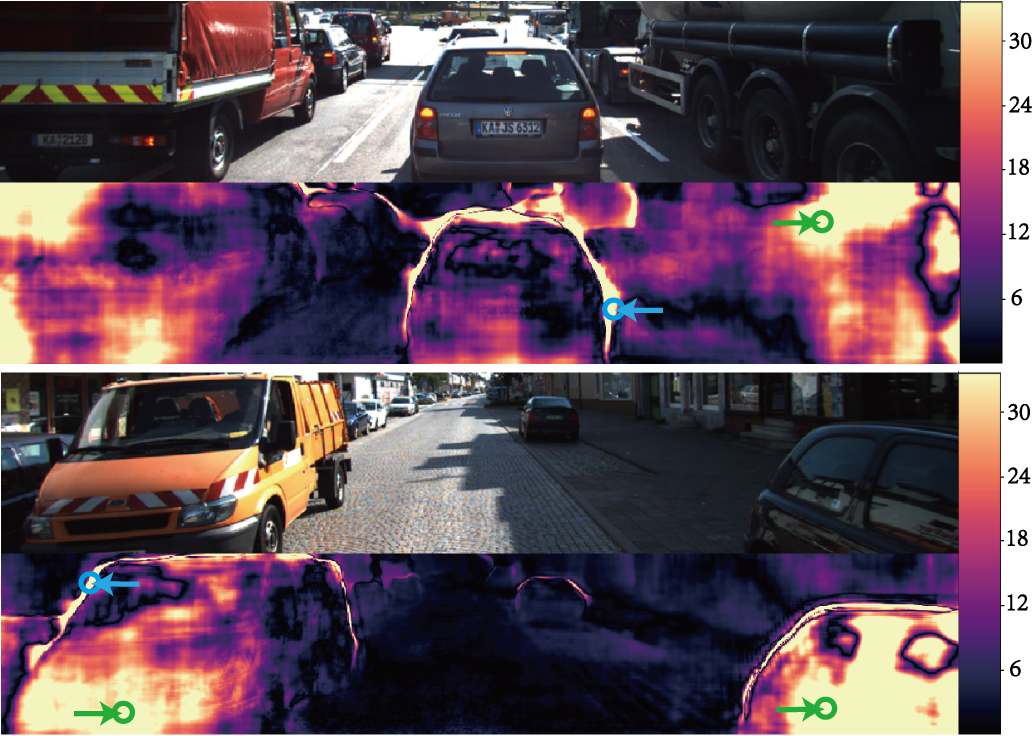}
   \vspace{-2mm}
 \caption{Input image and the disagreement of estimated disparity between our method and~\cite{watson2019self}. Our method impacts both borders (\textcolor{blue}{$\leftarrow$}) and inside (\textcolor{darkgreen}{$\rightarrow$}) of objects.}
 \label{fig:comparetoGodard}
 \figvspace
\end{figure}


\Paragraph{Depth Performance Analysis:}
Our method aims to explicitly constrain the estimated depth edges to become similar to segmentation counterparts. 
Yet, we observe that the improvements to the depth estimation, while being emphasised near edges, are distributed in {\it more} spatial regions. 
To understand this effect, we look at three perspectives. 

Firstly, we demonstrate that depth performance is the most challenging near edges using the $\delta < 1.25$ metric. 
We consider a point $\mathbf{x}$ to be near an edge point $\mathbf{p}$ if below averaged edge consistence $l_c$, that is $\mid \mathbf{x} - \mathbf{p}\mid \leq 3$. \linebreak
We demonstrate the depth performance of off-edge, whole image, and near edge regions in Tab.~\ref{tab:edge_performance}. 
Although our method has superior performance on whole, \textit{each} method degrades near an edge ($\downarrow\Sim0.18$ on $\delta$ from W to N), reaffirming the challenge of depth around object boundaries. 

\begin{table*}
    \centering
    \resizebox{0.8\linewidth}{!}{%
    \begin{tabular}{|c|c|c|l|l|l|l|l|l|l|}
        \hline
         Category & Method  & Morph  &\cellcolor[RGB]{222, 164, 151}Abs Rel &\cellcolor[RGB]{222, 164, 151} Sq Rel &\cellcolor[RGB]{222, 164, 151} RMSE &\cellcolor[RGB]{222, 164, 151} RMSE log & \cellcolor[RGB]{155, 187, 228}$\delta  < 1.25 $ & \cellcolor[RGB]{155, 187, 228}$\delta  < 1.25^2 $ & \cellcolor[RGB]{155, 187, 228}$\delta  < 1.25^3 $\\
          \hline
          \multirow{2}{*}{Unsupervised}&\multirow{2}{*}{Watson \textit{et al}.\cite{watson2019self}} &$\color[RGB]{184, 37, 25}\text{\ding{55}}$&   $0.097$  &   $0.734$  &   $4.454$  &   $0.187$  &   $0.889$  &   $0.961$  &   $0.981$ \\
          & & $\color[RGB]{66, 146, 41}\checkmark$ &   $0.096\color[RGB]{66, 146, 41}\downarrow$  &   $0.700\color[RGB]{66, 146, 41}\downarrow$  &   $4.401\color[RGB]{66, 146, 41}\downarrow$  &   $0.184\color[RGB]{66, 146, 41}\downarrow$  &   $0.891\color[RGB]{66, 146, 41}\uparrow$  &   $0.963\color[RGB]{66, 146, 41}\uparrow$  &   $0.982\color[RGB]{66, 146, 41}\uparrow$   \\
         \hline
          \multirow{2}{*}{Supervised}& \multirow{2}{*}{Lee \textit{et al}. \cite{lee2019big}} &$\color[RGB]{184, 37, 25}\text{\ding{55}}$&   $0.088$  &   $0.490$  &   $3.677$  &   $0.168$  &   $0.913$  &   $0.969$  &   $0.984$ \\
          && $\color[RGB]{66, 146, 41}\checkmark$&   $0.088$  &   $0.488\color[RGB]{66, 146, 41}\downarrow $  &   $3.666\color[RGB]{66, 146, 41}\downarrow$  &   $0.168$  &   $0.913$  &   $0.970\color[RGB]{66, 146, 41}\uparrow$  &   $0.985\color[RGB]{66, 146, 41}\uparrow$ \\
          \hline
          \multirow{2}{*}{Stereo}&\multirow{2}{*}{Yin \textit{et al}.\cite{yin2019hierarchical}} &$\color[RGB]{184, 37, 25}\text{\ding{55}}$&   $0.049$  &   $0.366$  &   $3.283$  &   $0.153$  &   $0.948$  &   $0.971$  &   $0.983$\\
          & & $\color[RGB]{66, 146, 41}\checkmark$ &   $0.049$  &   $0.365\color[RGB]{66, 146, 41}\downarrow$  &   $3.254\color[RGB]{66, 146, 41}\downarrow$  &   $0.152\color[RGB]{66, 146, 41}\downarrow$  &   $0.948$  &   $0.971$  &   $0.983$ \\
          \hline
    \end{tabular}
    }
    \vspace{-2mm}
    \caption{Comparison of algorithms if coupled with an segmentation network during inference. Given the segmentation predicted at inference, we apply morph defined in Sec.~\ref{sec:depthmorphing} to depth prediction. The improved metric is marked in green.}
    \label{tab:morph}
    \figvspace
    \vspace{2mm}
\end{table*}

\begin{table*}[ht]
    \centering
    \resizebox{0.8\linewidth}{!}{%
    \begin{tabular}{|c|c|c|c|c|c|c|c|c|}
        \hline
         Loss & Morph &\cellcolor[RGB]{222, 164, 151}Abs Rel &\cellcolor[RGB]{222, 164, 151} Sq Rel &\cellcolor[RGB]{222, 164, 151} RMSE &\cellcolor[RGB]{222, 164, 151} RMSE log & \cellcolor[RGB]{155, 187, 228}$\delta  < 1.25 $ & \cellcolor[RGB]{155, 187, 228}$\delta  < 1.25^2 $ & \cellcolor[RGB]{155, 187, 228}$\delta  < 1.25^3 $\\
         \hline
         Baseline & $\color[RGB]{184, 37, 25}\text{\ding{55}}$ &   $0.102$  &   $0.754$  &   $4.499$  &   $0.187$  &   $0.884$  &   $0.962$  &   $0.982$ \\
         \hline
         Baseline $+$ $\mathbf{M}$ &  $\color[RGB]{184, 37, 25}\text{\ding{55}}$ & $0.101$  &   $0.762$  &   $4.489$  &   $0.186$  &   $0.887$  &   $0.962$  &   $0.982$ \\
         \hline
         \multirow{2}{*}{Baseline $+$ $\mathbf{M}$ $+$  $l_g$} &  $\color[RGB]{184, 37, 25}\text{\ding{55}}$ &   $0.099$  &   $0.736$  &   $4.462$  &   $0.185$  &   $0.889$  &   $0.963$  &   $0.982$ \\
         & $\color[RGB]{66, 146, 41}\checkmark$ &   $0.098$  &   $0.714$  &   $4.421$  &  $0.183$  &   $0.890$  &  $ 0.964$  &   $0.982$    \\
         \hline
         \multirow{2}{*}{Baseline $+$ $\mathbf{M}$ $+$ $l_g$ $+$ $\text{Finetune}$} &  $\color[RGB]{184, 37, 25}\text{\ding{55}}$ &   $0.098$  &   $0.692$  &   $4.393$  &   $0.182$  &   $0.889$ &   $0.963$  &   $\mathbf{0.983}$ \\
         &$\color[RGB]{66, 146, 41}\checkmark$&    $\mathbf{0.097}$  &   $\mathbf{0.674}$  &   $\mathbf{4.354}$  &   $\mathbf{0.180}$  &   $\mathbf{0.891}$  &   $\mathbf{0.964}$  &   $\mathbf{0.983}$  \\
        \hline
    \end{tabular}
    }
    \vspace{-2mm}
    \caption{Ablation study of the proposed method. $\color[RGB]{66, 146, 41}\checkmark$ indicates morphing during inference.}
    \label{tab:ablation}
    \figvspace
\end{table*}

\begin{figure}[t!]
  \includegraphics[width=0.99\linewidth]{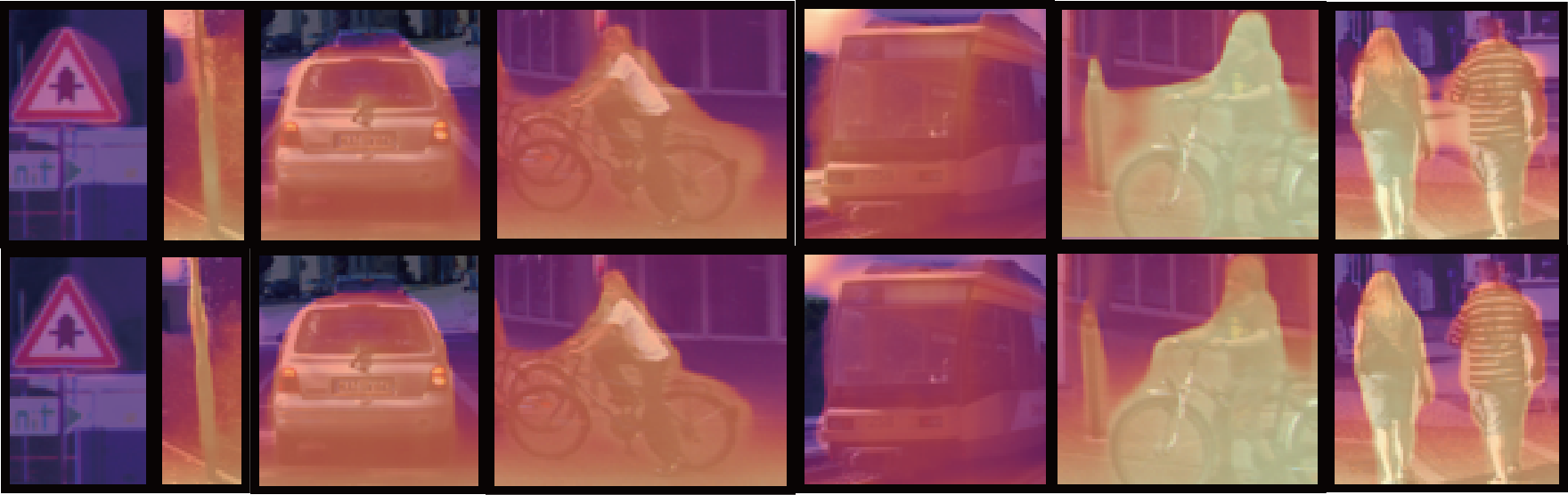}
  \vspace{-2mm}
  \caption{Compare the  quality of estimated depth around foreground objects between~\cite{watson2019self} (top) and ours (bottom).}
  \label{fig:qualitative_comparison}
  \figvspace
   \vspace{1mm}
\end{figure}

\begin{figure}[t!]
    \centering
    \includegraphics[width=\linewidth]{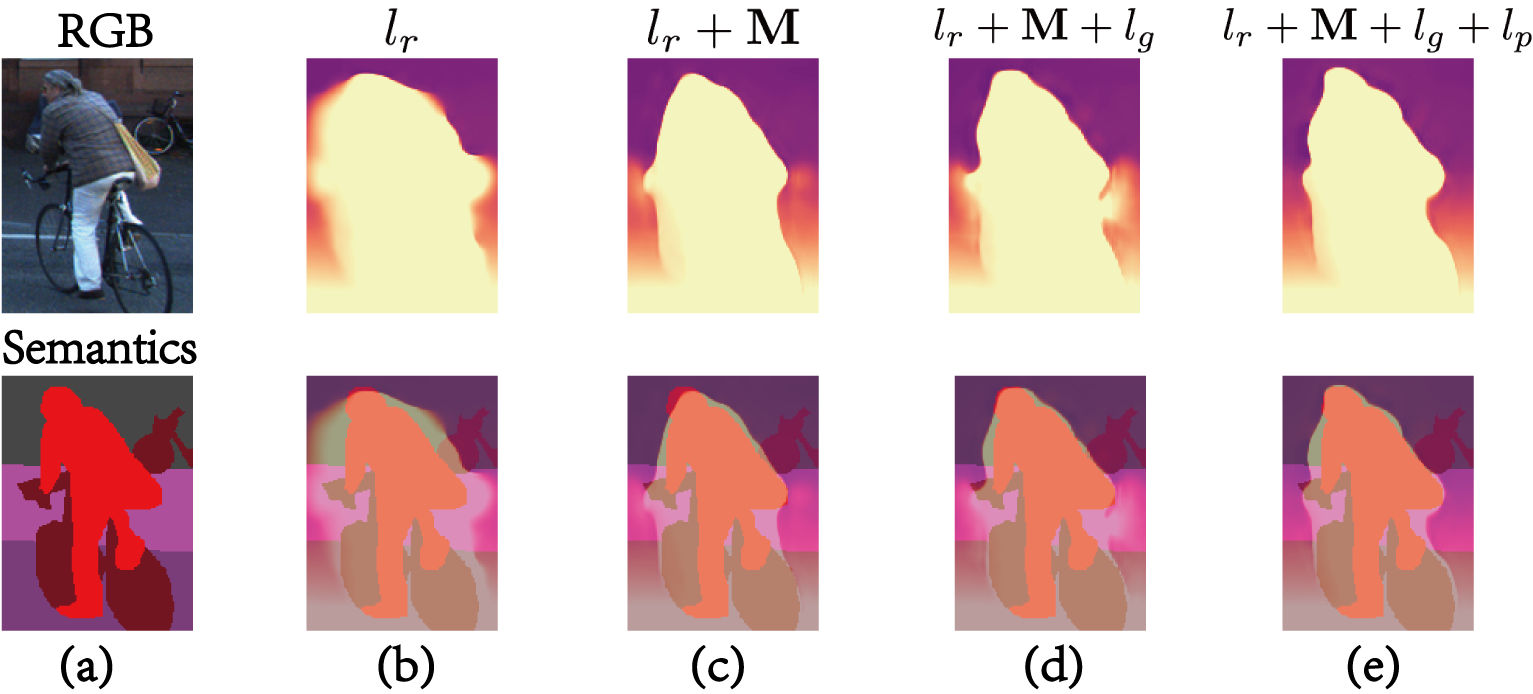}
    \vspace{-6mm}
    \caption{(a) input image and segmentation, (b-e) estimated depth (top) and with overlaid segmentation (bottom) for various ablation settings, as defined in Tab.~\ref{tab:ablation}.} 
    \label{fig:ablation_qualitive}
    \figvspace
    \vspace{-2mm}
\end{figure}

Secondly, we compare metric $\delta < 1.25$ against baseline~\cite{watson2019self} in the left axes of Fig.~\ref{fig:delta_funciton_of_pixel_distnace}. 
We observe improvement from background around object borders~($\text{px} \Sim -5$) and from foreground inside objects~($\text{px} \geq 30$). This is cross-validated in Fig.~\ref{fig:comparetoGodard} which visualizes the disagreements between ours and baseline~\cite{watson2019self}. Our method impacts near the borders (\textcolor{blue}{$\leftarrow$}) as well as inside of objects (\textcolor{darkgreen}{$\rightarrow$}) in Fig.~\ref{fig:comparetoGodard}. 

Thirdly, we view the improvement as a normalized probability distribution, as illustrated in right axes of Fig.~\ref{fig:delta_funciton_of_pixel_distnace}.\linebreak
It peaks at around $-5$ px, which agrees with the visuals of Fig.~\ref{fig:qualitative_comparison} where originally the depth spills into the background but becomes close to object borders using ours. 
Still, the improvement is consistently positive and generalized to entire distance range.
Such findings reaffirm that our improvement is both {\it near and beyond} the edges in a general manner.


\Paragraph{Depth Border Quality:}
We examine the quality of depth borders compared to the baseline~\cite{watson2019self}, as in Fig.~\ref{fig:qualitative_comparison}. \linebreak 
The depth borders of our proposed method is significantly more aligned to object boundaries.
We further show that for SOTA methods, even without training our models, applying our morphing step at inference leads to performance gain,
when coupled with a segmentation network~\cite{zhu2019improving} (trained with only $200$ domain images).  
As in Tab.~\ref{tab:morph}, this trend holds for unsupervised, supervised, and multi-view depth inference systems, implying that typical depth methods can struggle with borders, where our morphing can augment. 
However, we find that the inverse relationship using depth edges to morph segmentation is harmful to border quality. 

\Paragraph{Stereo Occlusion Mask:}
To examine the effect of our proposed stereo occlusion masking (Sec.~\ref{sec:occlusion_masking}), we ablate its effects (Tab.~\ref{tab:ablation}). 
The stereo occlusion mask $\mathbf{M}$ improves the absolute relative error ($0.102 \to 0.101$) and $\delta < 1.25$ ($0.884 \to 0.887$).
Upon applying stereo occlusion mask during training, we observe the bleeding artifacts are significantly controlled as in Fig.~\ref{fig:ablation_qualitive} and in \textbf{Suppl.}~Fig.~$3$.
Hence, the resultant borders are stronger, further supporting the proposed consistency  term $l_c$ and  morphing operation.

\Paragraph{Morph Stabilization:}
We utilize estimated segmentation~\cite{zhu2019improving} to define the segmentation-depth edge morph. 
Such estimations inherently introduce noise and destablization in training for which we propose a $\mathbf{w}(\mathbf{x})$ weight to provide less attention to low image variance and ignore any regions which degrades photometric loss (Sec.~\ref{sec:lossfunction}). 
Additionally, we ablate the specific help from stereo proxy labels in stabilizing training in Fig.~\ref{fig:ablation_qualitive} (d) \& (e) and \textbf{Suppl.}~Fig.~$3$.

\begin{table}[t!]
    \centering
    \setlength\tabcolsep{1pt}
    \resizebox{\linewidth}{!}{
    \begin{tabular}{|c|c|c|c|c|c|c|}
        \hline
         Model & Finetune &\cellcolor[RGB]{222, 164, 151}Abs Rel &\cellcolor[RGB]{222, 164, 151} Sq Rel &\cellcolor[RGB]{222, 164, 151} RMSE &\cellcolor[RGB]{222, 164, 151} RMSE log &
         \cellcolor[RGB]{155, 187, 228}$\delta  < 1.25 $ \\
        \hline
         \multirow{2}{*}{Godard \textit{et al}.~\cite{godard2019digging}} &$\color[RGB]{184, 37, 25}\text{\ding{55}}$ &   $0.104$  &   $0.775$  &   $4.562$  &   $0.191$  &  $0.878$  \\
        &$\color[RGB]{66, 146, 41}\checkmark$ &   $0.103$  &   $0.731$  &   $4.531$  &   $0.188$  &  $0.878$  \\
        \hline
    \multirow{2}{*}{Watson \textit{et al}.~\cite{watson2019self}} &$\color[RGB]{184, 37, 25}\text{\ding{55}}$&   $0.096$  &   $0.710$  &   $4.393$  &   $0.185$  &  $0.890$   \\
        &$\color[RGB]{66, 146, 41}\checkmark$  &   $0.094$  &   $0.676$  &   $4.317$  &   $0.180$  &  $0.892$    \\
        \hline
    \end{tabular}}
    \vspace{-2mm}
    \caption{Improvement after finetuning of different models.}
    \label{tab:finetune}
    \figvspace
     \vspace{1mm}
\end{table}

\Paragraph{Finetuning Strategy:}
To better understand the effect of our finetuning strategy (Sec.~\ref{sec:lossfunction}) on performance, we ablate using ~\cite{godard2019digging, watson2019self} and our method, as shown in Tab.~\ref{tab:ablation} and \ref{tab:finetune}.\linebreak 
Each ablated method achieves better performance after applying the finetuning, suggesting the technique is general. 

\begin{figure}
  \centering
  \includegraphics[width=0.75\linewidth]{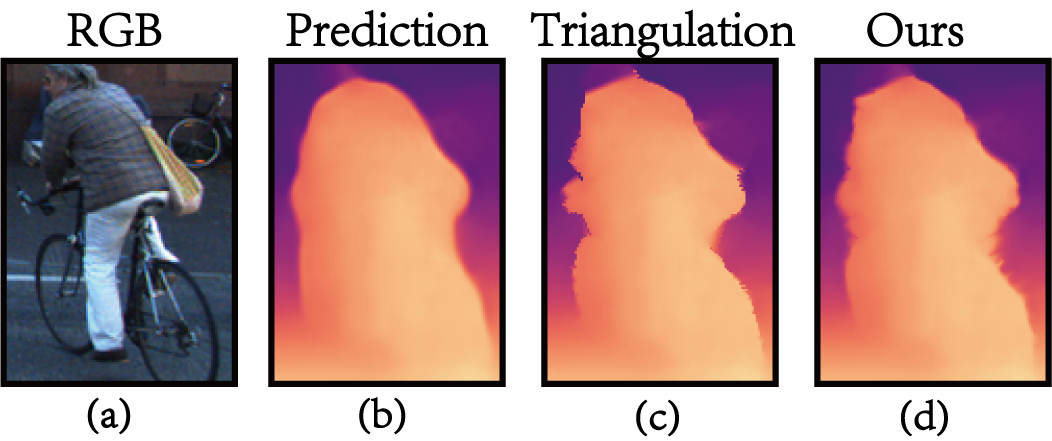}
  \vspace{-2mm}
  \caption{ Comparison of depth of initial baseline (b), triangularization (c), and proposed morph (d).}
  \label{fig:morph_ablation}
  \figvspace
\end{figure}

\Paragraph{Morphing Strategy:}
We explore the sensitivity of our morph operation (Sec.~\ref{sec:explicit_consistency_section}), by comparing its effectiveness against using triangularization to distill point pair relationships. 
We accomplish this by first forming a grid over the image using anchors.
Then define corresponding triangularization pairs between the segmentation edge points paired with two anchors. 
Lastly, we compute an affine transformation between the two triangularizations. 
We analyze the technique vs.~our proposed morphing strategy qualitatively in Fig.~\ref{fig:morph_ablation} and quantitatively in Tab.~\ref{tab:morph_compare}. 
Although the methods have subtle distinctions, the triangularization morph is generally inferior, as highlighted by the RMSE metrics in Tab.~\ref{tab:morph_compare}.
Further, the triangularization morphing forms boundary errors with acute angles which introduce more noise in the supervision signal, as exemplified in Fig.~\ref{fig:morph_ablation}.

\begin{table}[t!]
    \centering
    \small
    \setlength\tabcolsep{1pt}
    \resizebox{0.91\linewidth}{!}{
    \begin{tabular}{|c|c|c|c|c|}
        \hline
         Method  &\cellcolor[RGB]{222, 164, 151} Sq Rel &\cellcolor[RGB]{222, 164, 151} RMSE &\cellcolor[RGB]{222, 164, 151} RMSE log & \cellcolor[RGB]{155, 187, 228}$\delta  < 1.25 $\\
        \hline
        Ours (Triangularization)  &   $0.697$   &   $4.379$  &   $0.180$ & $0.895$  \\
        \hline
        Ours (Proposed)  &   $0.686$    &   $4.368$&   $0.180$ & $0.895$ \\
        \hline
    \end{tabular}
    }
    \vspace{-2mm}
    \caption{Our morphing strategy versus triangularization.}
    \label{tab:morph_compare}
    \figvspace
\end{table}

\Section{Conclusions}
We present a depth estimation framework designed to explicitly consider the mutual benefits between two neighboring computer vision tasks of self-supervised depth estimation and semantic segmentation. 
Prior works have primarily considered this relationship implicitly. In contrast, we propose a morphing operation between the borders of the predicted segmentation and depth, then use this morphed result as an additional supervising signal. 
To help the edge-edge consistency quality, we identify the source problem of bleeding artifacts near object boundaries then propose a stereo occlusion masking to alleviate it. 
Lastly, we propose a simple but effective finetuning strategy to further boost generalization performance. 
Collectively, our method advances the state of the art on self-supervised depth estimation, matching the capacity of supervised methods, and significantly improves the border quality of estimated depths. 

\Paragraph{Acknowledgment} 
Research was partially sponsored by the Army Research Office under Grant Number W911NF-18-1-0330. The views and conclusions contained in this document are those of the authors and should not be interpreted as representing the official policies, either expressed or implied, of the Army Research Office or the U.S.~Government. The U.S.~Government is authorized to reproduce and distribute reprints for Government purposes notwithstanding any copyright notation herein.

\FloatBarrier
{\small
\bibliographystyle{ieee_fullname}
\bibliography{ExplicitConstraintSegDepth} 
}

\pagebreak

{\LARGE \noindent \textbf{Supplementary Material} \par}
\setcounter{equation}{0}
\setcounter{figure}{0}
\setcounter{table}{0}
\setcounter{section}{0}
\section{Proof of Local Optimality}
\label{sec:proof}

 We give a brief proof that, under constructed transformation set $\{\phi(\mathbf{x} \mid \mathbf{q}, \mathbf{p})\}$, the proposed edge-edge consistency $l_c(\mathbf{\Gamma}(\mathbf{T_s} \mid \mathbf{T_d}),~ \mathbf{T^*_d})$, can achieve the local optimality when the segmentation-augmented (or morphed) disparity edge points satisfy $\mathbf{T^*_d} = \{\mathbf{p}\, \mid\,\norm{\frac{\partial \mathbf{I_{\hat{d}}^*}(\mathbf{p})}{\partial \mathbf{x}}} > \frac{t}{1 + t} \cdot k_1\}$. 

To prove this, let's start by evaluating the  gradient of morphed disparity map $\mathbf{I_{\hat{d}}^*}$ at a semantic edge pixel $\mathbf{q}$:
\begin{gather}
    \begin{aligned}
    \forall \mathbf{q} \in \mathbf{\Gamma}(\mathbf{T_s} \mid \mathbf{T_d}), \,\, \frac{\partial \mathbf{I_{\hat{d}}^*}(\mathbf{x})}{\partial \mathbf{x}} \bigg|_{\mathbf{x} = \mathbf{q}} &= \frac{\partial  \mathbf{I_{\hat{d}}}(\phi(\mathbf{x}))}{\partial \phi(\mathbf{x})} * \frac{\partial \phi(\mathbf{x})}{\partial \mathbf{x}} \bigg|_{\mathbf{x} = \mathbf{q}} \\
    & =\frac{\partial \mathbf{I_{\hat{d}}}(\mathbf{y})}{\partial \mathbf{y}} \bigg|_{\mathbf{y} = \mathbf{p}} *\frac{\partial \phi(\mathbf{x})}{\partial \mathbf{x}} \bigg|_{\mathbf{x} = \mathbf{q}}, \\
    \end{aligned}
    \label{eq:opt1}
\end{gather}
Note if $\mathbf{x} = \mathbf{q}$, $\phi(\mathbf{x} \mid \mathbf{q}, \mathbf{p}) = \mathbf{p}$. 
If $\frac{\partial \mathbf{I_{\hat{d}}^*}(\mathbf{x})}{\partial \mathbf{x}} \big|_{\mathbf{x} = \mathbf{q}}$ is sufficiently larger than a threshold, a semantic edge pixel $\mathbf{q}$ is also an edge pixel in the morphed disparity map, leading to the perfect edge-edge consistency for $\mathbf{q}$.
We now derive the two terms in Eq.~\ref{eq:opt1}, in order to find that threshold. 

When $\mathbf{x}$ is on the line segment $\protect\overrightarrow{\mathbf{qp}}$, its projection $\mathbf{x'}$ overlaps with itself. 
We can thus compute $\frac{\partial \phi(\mathbf{x})}{\partial \mathbf{x}} \big|_{\mathbf{x} = \mathbf{q}}$ as:
\begin{equation}
    \centering
    \begin{aligned}
    \frac{\partial \phi(\mathbf{x})}{\partial \mathbf{x}} \bigg|_{\mathbf{x} = \mathbf{q}}& = \frac{\partial \left(\mathbf{x} +  \protect\overrightarrow{\mathbf{qp}} - \frac{1}{1+t}\cdot \protect\overrightarrow{\mathbf{qx'}}\right) }{\partial \mathbf{x}} \bigg|_{\mathbf{x} = \mathbf{q}} \\
    &=\frac{\partial \left(\mathbf{x} +  \protect\overrightarrow{\mathbf{qp}} - \frac{1}{1+t}\cdot \protect\overrightarrow{\mathbf{qx}}\right) }{\partial \mathbf{x}} \bigg|_{\mathbf{x} = \mathbf{q}} \\
    &=\frac{\partial \left(\mathbf{x} + (\mathbf{p} - \mathbf{q}) - \frac{1}{1+t}\cdot (\mathbf{x} - \mathbf{q}) \right) }{\partial \mathbf{x}} \bigg|_{\mathbf{x} = \mathbf{q}}\\
    &=\frac{\partial \left(\frac{t}{1+t} \cdot \mathbf{x} + \mathbf{p} - \frac{t}{1+t}\cdot \mathbf{q} \right) }{\partial \mathbf{x}} \bigg|_{\mathbf{x} = \mathbf{q}}\\
    &= \frac{t}{1 + t}.
    \end{aligned}
\end{equation}
Using $\frac{\partial \phi(\mathbf{x})}{\partial \mathbf{x}} \big|_{\mathbf{x} = \mathbf{q}} = \frac{t}{1 + t}$ with Eq.~\ref{eq:opt1}, we have:
\begin{gather}
    \begin{aligned}
    \forall \mathbf{q} \in \mathbf{\Gamma}(\mathbf{T_s} \mid \mathbf{T_d}), \,\, \frac{\partial \mathbf{I_{\hat{d}}^*}(\mathbf{x})}{\partial \mathbf{x}} \bigg|_{\mathbf{x} = \mathbf{q}} & =\frac{\partial \mathbf{I_{\hat{d}}}(\mathbf{y})}{\partial \mathbf{y}} \bigg|_{\mathbf{y} = \mathbf{p}} *\frac{\partial \phi(\mathbf{x})}{\partial \mathbf{x}} \bigg|_{\mathbf{x} = \mathbf{q}} \\
    &= \frac{t }{1 + t} * \frac{\partial \mathbf{I_{\hat{d}}}(\mathbf{y})}{\partial \mathbf{y}} \bigg|_{\mathbf{y} = \mathbf{p}}\\ 
    &> \frac{t }{1 + t} \cdot k_1,
    \end{aligned}
    \label{eq:opt2}
\end{gather}
where the inequality is derived from Eq.~1 of the main paper, which defines the threshold $k_1$ for detecting edge pixels on the original disparity map.
Here, in morphed disparity map $\mathbf{I_{\hat{d}}^*}$, since every counted semantic edge pixel $\mathbf{q} \in \mathbf{\Gamma}(\mathbf{T_s} \mid \mathbf{T_d})$ in computing the consistency $l_c$ has a gradient magnitude larger than the threshold $\frac{t }{1 + t} \cdot k_1$, $\mathbf{q}$ overlaps with the paired or matched depth/disparity edge pixel $\mathbf{p}$ as well, 
{\it i.e.}, $\mathbf{T^*_d} = \{\mathbf{p}\, \mid\,\norm{\frac{\partial \mathbf{I_{\hat{d}}^*}(\mathbf{p})}{\partial \mathbf{x}}} > \frac{t}{1 + t} \cdot k_1\}$. Thus, in morphed disparity map $\mathbf{I_{\hat{d}}^*}$, semantic border overlaps with depth borders, making proposed consistency measurement $l_c$ hit local minimum $0$:
\begin{gather}
    \begin{aligned}
    &\forall \mathbf{q} \in \mathbf{\Gamma}(\mathbf{T_s} \mid \mathbf{T_d}), \,\, \frac{\partial \mathbf{I_{\hat{d}}^*}(\mathbf{x})}{\partial \mathbf{x}} \bigg|_{\mathbf{x} = \mathbf{q}} > \frac{t }{1 + t} \cdot k_1 \\
    \iff &\forall \mathbf{q} \in \mathbf{\Gamma}(\mathbf{T_s} \mid \mathbf{T_d}), \,\,\,\,\,\, \delta(\mathbf{q}, \mathbf{T_d^*}) = \min_{\{\mathbf{p} \in \mathbf{T_d^*}\}}\norm{\mathbf{p} - \mathbf{q}}\\
    &{} \qquad \qquad \qquad \qquad \qquad \qquad \,\,\,= \norm{\mathbf{q} - \mathbf{q}} = 0\\
    \iff &l_c(\mathbf{\Gamma}(\mathbf{T_s} \mid \mathbf{T_d}), \,\mathbf{I_d^*}) = 0.\qquad \,\,\,{}
    \end{aligned}
    \label{eq:opt3}
\end{gather}
This shows that, under the defined transformation, we are realigning the depth edge set $\mathbf{\Omega}$ to the segmentation edge set $\mathbf{\Gamma^d_s}$, making the edge-edge consistency a local optimality.

Note that the threshold $\frac{t }{1 + t} \cdot k_1$ is not actually being applied to the morphed disparity map for  edge detection. Rather, we derive it as the condition that will be naturally satisfied in our work, when both the morph function and $k_1$ threshold for disparity map depth estimation (Eq.~1 of the main paper) are employed.

\begin{table}[]
    \centering
    \small
    \setlength\tabcolsep{4pt}
    \begin{tabular}{|l|l|l|l|l|l|l|}
        \hline
        \multicolumn{7}{|l|}{\textbf{Depth Decoder}}  \\
        \hline
        \textbf{layer} & \textbf{k} & \textbf{s} & \textbf{c} & \textbf{res} & \textbf{input} & \textbf{activation} \\
        \hline
        $\text{upconv5}$ & $3$ & $1$ & $256$ & $32$ & $\text{econv5}$ & $\text{ELU}$\cite{clevert2015fast} \\
        $\text{iconv5}$ & $3$ & $1$ & $256$ & $16$ & $\uparrow \text{upconv5, econv4}$ & $\text{ELU}$ \\
        \hline
        $\text{upconv4}$ & $3$ & $1$ & $128$ & $16$ & $\text{iconv5}$ & $\text{ELU}$ \\
        $\text{iconv4}$ & $3$ & $1$ & $128$ & $8$ & $\uparrow \text{upconv4, econv3}$ & $\text{ELU}$ \\
        $\text{disp4}$ & $3$ & $1$ & $1$ & $1$ & $\text{iconv4}$ & $\text{Sigmoid}$ \\
        \hline
        $\text{upconv3}$ & $3$ & $1$ & $64$ & $8$ & $\text{iconv4}$ & $\text{ELU}$ \\
        $\text{iconv3}$ & $3$ & $1$ & $64$ & $4$ & $\uparrow \text{upconv3, econv2}$ & $\text{ELU}$ \\
        $\text{disp3}$ & $3$ & $1$ & $1$ & $1$ & $\text{iconv3}$ & $\text{Sigmoid}$ \\
        \hline
        $\text{upconv2}$ & $3$ & $1$ & $32$ & $4$ & $\text{iconv3}$ & $\text{ELU}$ \\
        $\text{iconv2}$ & $3$ & $1$ & $32$ & $2$ & $\uparrow \text{upconv2, econv1}$ & $\text{ELU}$ \\
        $\text{disp2}$ & $3$ & $1$ & $1$ & $1$ & $\text{iconv2}$ & $\text{Sigmoid}$ \\ 
        \hline
        $\text{upconv1}$ & $3$ & $1$ & $16$ & $2$ & $\text{iconv2}$ & $\text{ELU}$ \\
        $\text{iconv1}$ & $3$ & $1$ & $16$ & $1$ & $\uparrow \text{upconv1}$ & $\text{ELU}$ \\
        $\text{disp1}$ & $3$ & $1$ & $1$ & $1$ & $\text{iconv1}$ & $\text{Sigmoid}$ \\
        \hline
    \end{tabular}
    \caption{The network architecture of our decoder. \textbf{k}, \textbf{s} and \textbf{c} denote the kernel size, stride and output channel numbers of the layer, respectively. \textbf{res} refers to relative downsampling scale to  the input image. $\uparrow$ symbol means a $2 \times$ nearest-neighbour upsampling to input.}
    \label{tab:decoder}\vspace{-4mm}
\end{table}
\section{Network details}
Across our experiments, we use ImageNet~\cite{deng2009imagenet} pretrained ResNet18 and ResNet50~\cite{he2016deep} as our encoder. Our decoder structure is same as Godard~\textit{et al.}~\cite{godard2019digging} and Waston~\textit{et al.}~\cite{watson2019self}, as detailed in Table~\ref{tab:decoder}. 
We also incorporate other practices such as color augmentation, random flip, edge-aware smoothness and exclusion of stationary pixels.
\section{More Ablations}\label{sec:supp_aba}
In this section, we perform additional ablations to further validate our proposed approach. 
We ablate (1) Our proposed morph strategy achieves local optimality of edge-edge consistency $l_c$, and (2) The stereo occlusion mask $\mathbf{M}$ boosts clear borders.
All our ablations are conducted on Eigen~\cite{eigen2014depth} test splits of KITTI~\cite{geigerwe}.

\begin{figure}
    \centering
    \vspace{-2mm}
    \includegraphics[width = 0.9\linewidth]{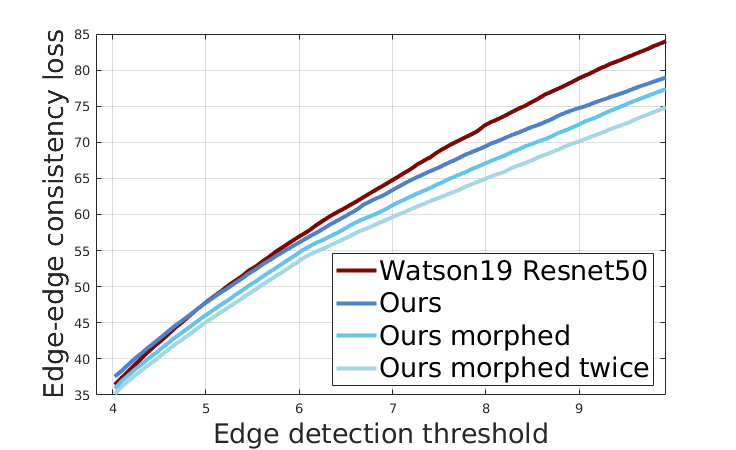}
    \vspace{-1mm}
    \caption{
    We plot the edge-edge consistency $l_c$ between Watson19~\cite{watson2019self} and ours at different edge detection thresholds  $k_1$. Additionally, we show the change of consistency $l_c$ after applying morph strategy once and twice during inference, in addition to using our learned network.
    }
    \label{fig:supp_aba_morph}\vspace{-4mm}
\end{figure}

\Paragraph{Reducing edge-edge consistency via morphing:}
We plot the edge-edge consistency loss $l_c$  under various edge detection thresholds $k_1$ in Fig.~\ref{fig:supp_aba_morph}. 
We cross-validate morphing~(detailed in main paper Section~$\textbf{3.1}$) as a technique to achieve local optimality of $l_c$ from Fig.~\ref{fig:supp_aba_morph} via showing  consistently decreased measurement $l_c$ after applying morphing once and twice. 
The lower loss in Fig.~\ref{fig:supp_aba_morph} shows that our models are more consistent with segmentation compared to~\cite{watson2019self}.
Additionally, increased threshold $k_1$ leads to thinner edges and neglects distant objects, which have two effects. 
First of all, thinner edges make edge-edge consistency to be more challenge, thus higher loss values.
Second, focusing on close-range objects can best leverage the high-quality segmentation, which leads to larger improvement margin over the baseline~\cite{watson2019self}. 


\begin{figure}
    \centering
    \includegraphics[width = 0.8\linewidth]{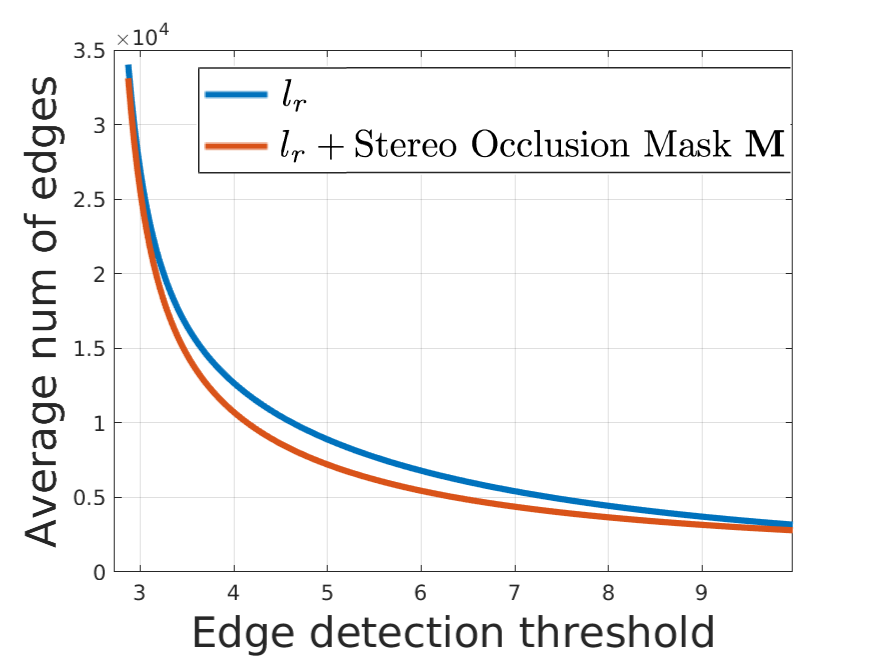}
    \vspace{-2mm}
    \caption{The effects of proposed stereo occlusion mask $\mathbf{M}$. We plot the trend of the average detected edge numbers $\frac{1}{n}\sum_{i = 1}^{i = n}(\|\frac{\partial \mathbf{I_{\hat{d}}^i}(\mathbf{x})}{\partial \mathbf{x}}\| > k_1)$ at different edge detection thresholds $k_1$, where n is for total number of tested images.}
    \label{fig:supp_aba_mask}\vspace{-2mm}
\end{figure}

\begin{figure*}
    \centering
    \includegraphics[width = \linewidth]{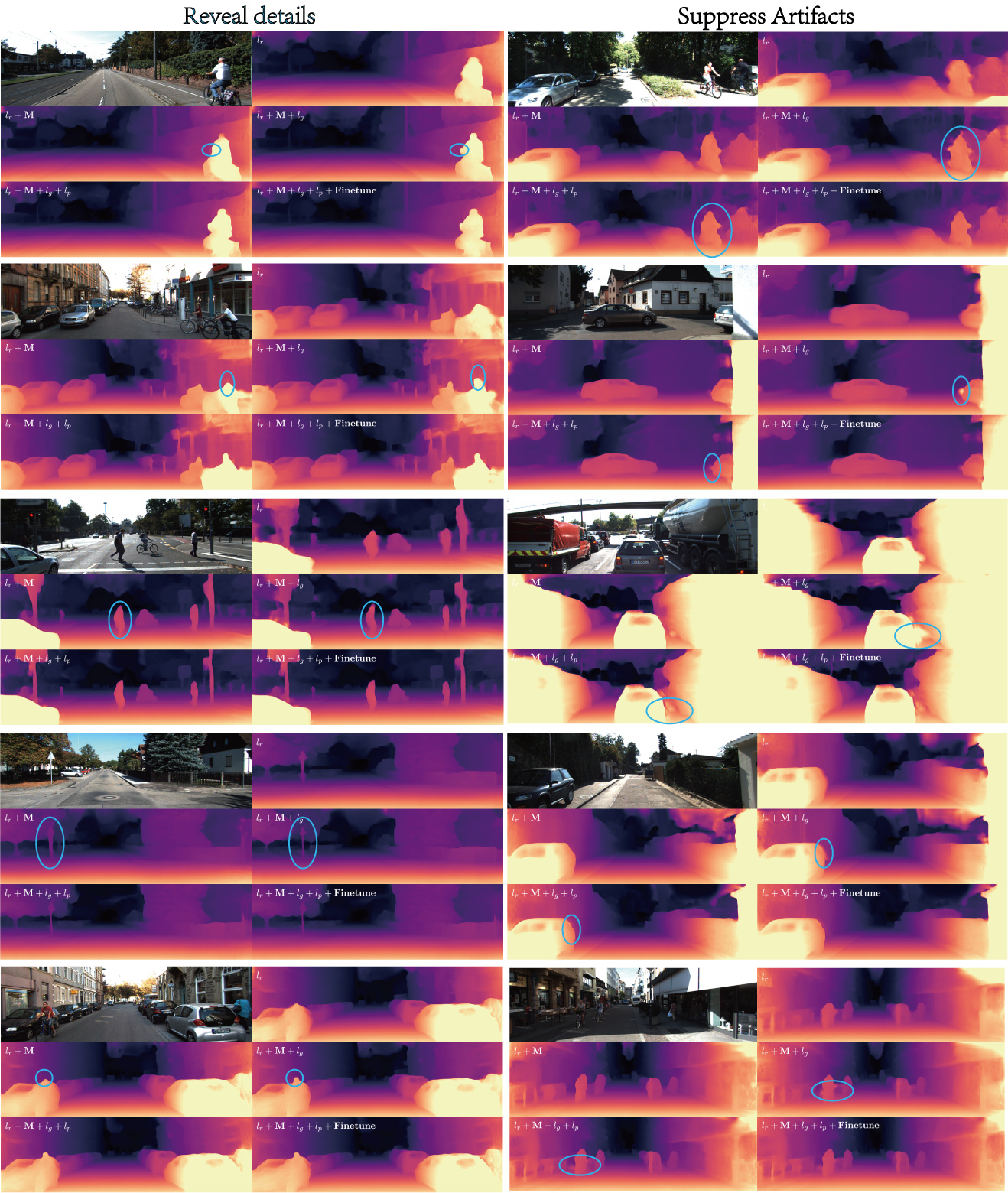}
    \caption{On the left column, explicit utilization of segmentation information helps recovering more details. On the the right, we show blobbed border artifacts in the low texture areas, caused by noisy predicted segmentation labels and low constrain from the photometric loss $l_r$. We suppress the artifacts by the incorporation of texture weight $\mathbf{w}$ and utilization of proxy stereo labels~\cite{kuznietsov2017semi, watson2019self}.}
    \label{fig:aba_aba}
\end{figure*}

\begin{figure*}
    \centering
    \includegraphics[width = \linewidth]{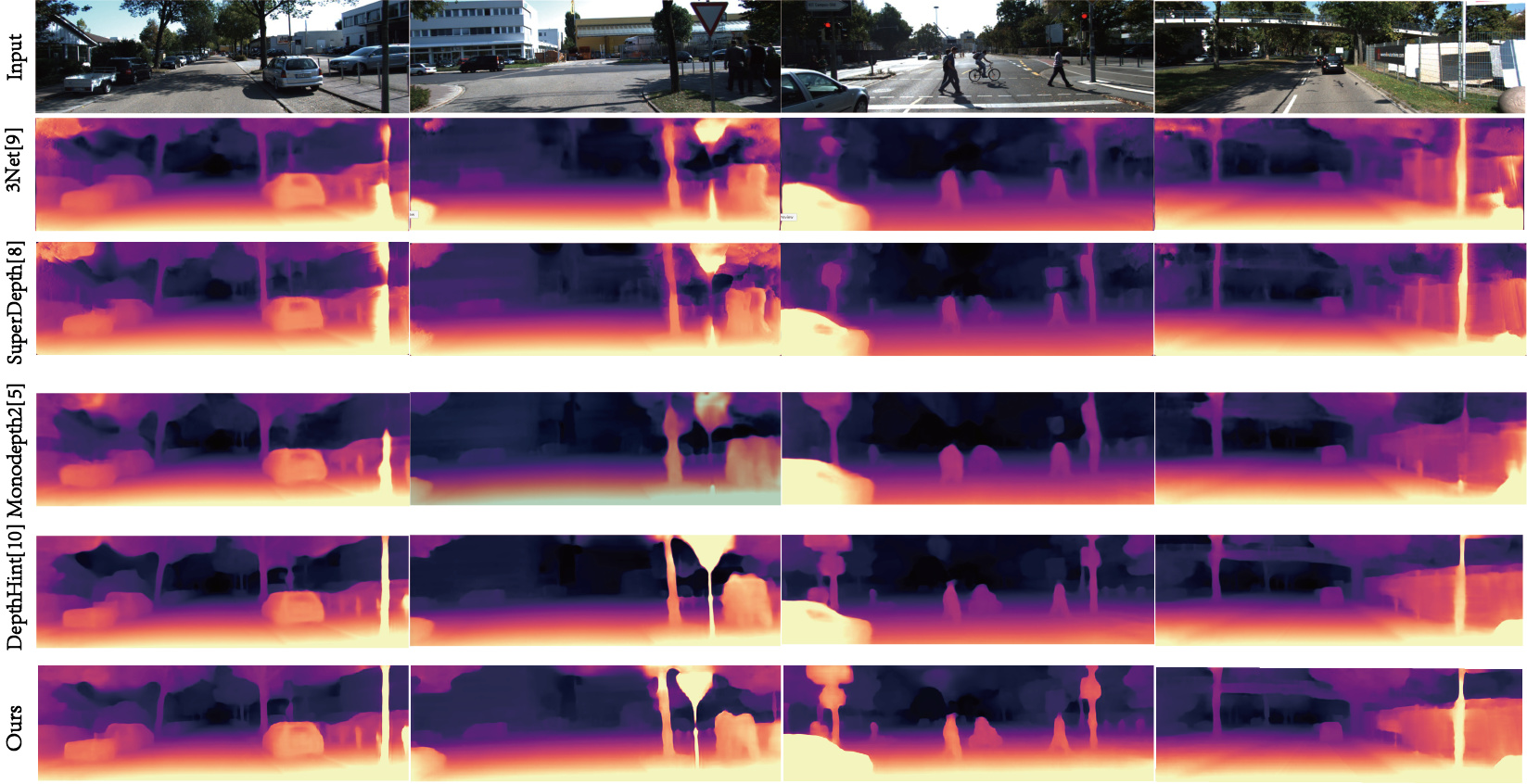}
    \caption{Comparison against other state of the arts~\cite{godard2019digging, pillai2019superdepth, poggi2018learning, watson2019self}. Our method reconstructs more object details compared to previous works and possesses the most clear border overall.}
    \vspace*{35in}
    \label{fig:aba_qua_compare}
\end{figure*}

\Paragraph{Stereo Occlusion Mask:}
In Fig.~\ref{fig:aba_qua_compare}, we observe bleeding artifacts universally exist in stereo-based systems~\cite{pillai2019superdepth, poggi2018learning, watson2019self}. 
In~\cite{watson2019self}, the utilization of stereo proxy label partially suppresses it as its additional constrain on the low texture area. 
~\cite{godard2019digging} reduces the artifacts via supervision from videos. 
In comparison, without any additional supervision sources, we eliminate it via the proposed stereo occlusion mask $\mathbf{M}$. As an example, the top-right subfigure of Fig.~\ref{fig:aba_aba} reveals a clearer and thinner border when comparing $l_r$ against $l_r + \mathbf{M}$.
This motivates us to treat ``thinness" as a measurement and use the average detected edge number $\frac{1}{n}\sum_{i = 1}^{i = n}(\|\frac{\partial \mathbf{I_{\hat{d}}^i}(\mathbf{x})}{\partial \mathbf{x}}\| > k_1)$ as an approximated metric of border clearance, as shown in Fig.~\ref{fig:supp_aba_mask}. 
As expected, after applying the mask $\mathbf{M}$, edges become more ``thinner" and clearer, reflected as the decreased number of detected edges.

\begin{figure*}
    \centering
    \includegraphics[width = \linewidth]{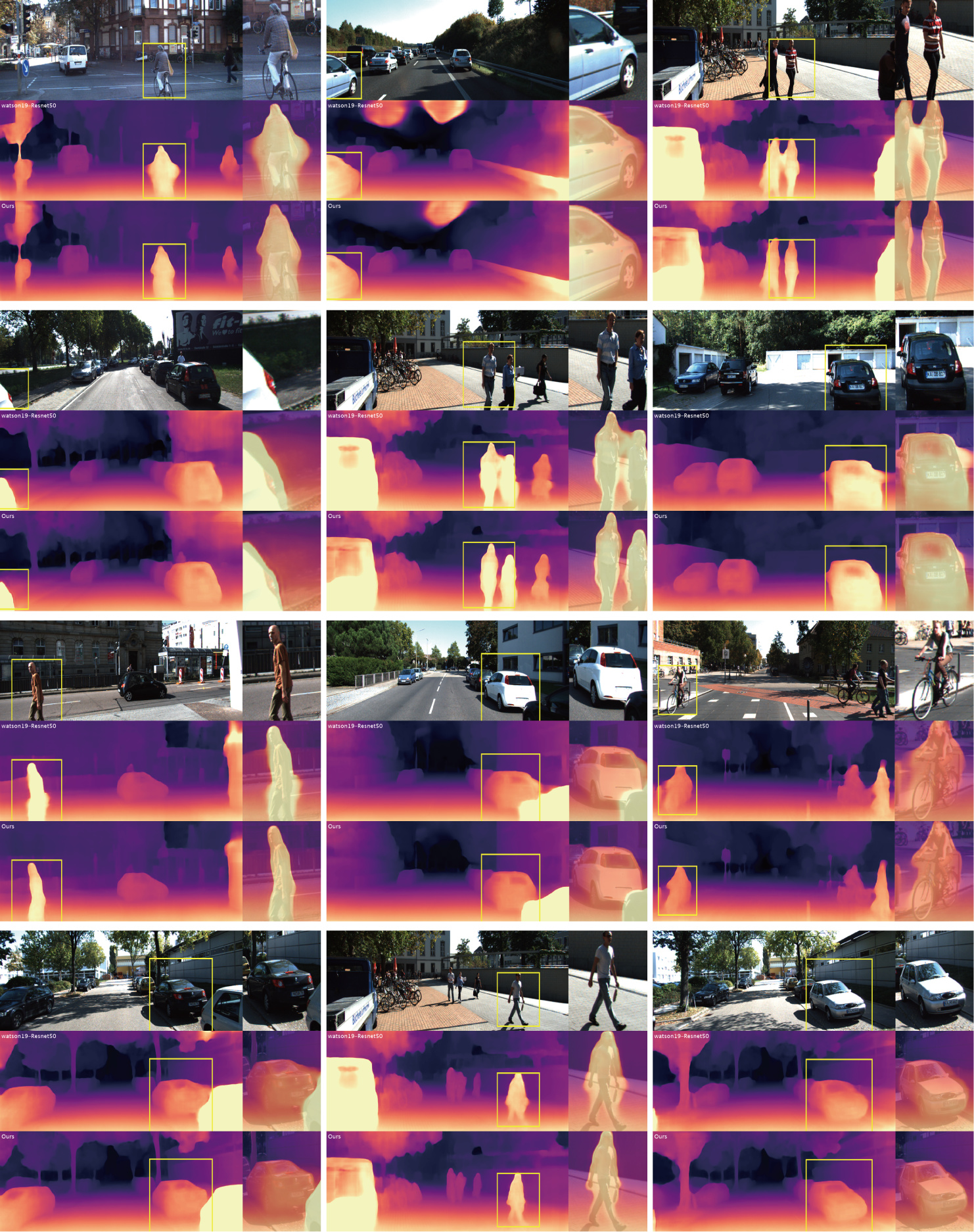}
    \caption{More comparison between ours model and the state-of-the-art baseline~\cite{watson2019self}. Content within yellow box is zoomed in and attached to the right. We show significantly improved border quality compared to the method of~\cite{watson2019self}.}
    \label{fig:aba_compare}
\end{figure*}

\Paragraph{More quality comparisons:}
We show additional qualitative examples when different loss are applied in Fig.~\ref{fig:aba_aba}. 
We further provide qualitative comparisons against the baseline method~\cite{watson2019self} in Fig.~\ref{fig:aba_compare}, and other methods in Fig.~\ref{fig:aba_qua_compare}.

\end{document}